\documentclass[twocolumn]{IEEEtran}

\usepackage[pdftex]{graphicx} % 插图命令 \includegraphics
\usepackage{amsmath} % Math Packages
\usepackage{algorithmic}
\usepackage{array} % Alignment Packages
\usepackage[caption=false,font=footnotesize]{subfig} % Subfigure Packages
\usepackage{dblfloatfix}
\usepackage{url} % simple URL typesetting

\usepackage{microtype}
\usepackage[utf8]{inputenc} % allow utf-8 input
\usepackage[T1]{fontenc}    % use 8-bit T1 fonts
\usepackage{paralist} % Compacted versions of enumerate and itemize.
\usepackage{marvosym} % 常用符号, 特别是打勾、打叉
\usepackage{bm} % 实现字体的斜体加粗
\usepackage{bbm} % fix \mathbb doesn't support digits
\usepackage{amssymb}   % 黑板粗体 \mathbb, 德文尖角体 \mathfrak, \backsim
\usepackage{mathrsfs} % 花体 \mathscr
\usepackage{tabularray} % tblr or talltblr or longtblr environment
\UseTblrLibrary{booktabs}       % professional-quality tables，也就是三线表
\usepackage{tabularx} % 自动计算列宽的表格包
\usepackage[flushleft]{threeparttable} % 给表格项添加注释
\usepackage{colortbl} % 表格行列颜色
\usepackage{makecell} % 表格内容换行
\usepackage{multirow} % 多行表格
\usepackage{nicefrac} % 斜着的分号
\usepackage[dvipsnames]{xcolor} % 字体颜色
\usepackage{soul} % highlight, strikeout, underline
\setul{0.5ex}{0.3ex}

\definecolor{citecolor}{RGB}{34,139,34}
\usepackage[pagebackref=false,breaklinks=true,letterpaper=true,colorlinks,
    citecolor=citecolor,bookmarks=false]{hyperref}
\usepackage{cleveref}  % for \cref
\crefformat{figure}{Fig.~#2#1#3}
\crefformat{equation}{Eq.~(#2#1#3)}
\crefformat{table}{Table~#2#1#3}
\crefformat{section}{Section~#2#1#3}
\crefformat{algorithm}{Algorithm~#2#1#3}
\crefformat{appendix}{Appendix #2#1#3}
\crefformat{subsection}{subsection~#2#1#3}
\usepackage{cellspace}
\usepackage[numbers,sort&compress]{natbib} % 代替 cite.sty

\usepackage{xspace} % 用于在 \newcommand 中添加空格

\usepackage{pifont} % http://ctan.org/pkg/pifont

\usepackage{soul} % \hl 高亮
\setul{0.3ex}{0.1pt} % 设置下划线的高度和粗细
\definecolor{modifiedblue}{RGB}{64,120,192}
\definecolor{Gray}{rgb}{0.9,0.9,0.9}
\definecolor{LightCyan}{rgb}{0.88,1,1}
\newcolumntype{a}{>{\columncolor{Gray}}c}
\newcolumntype{b}{>{\columncolor{white}}c}

\begin{document}

\setlength{\abovedisplayskip}{.5\baselineskip} % 调整公式与正文间的段前距离
\setlength{\belowdisplayskip}{.5\baselineskip} % 调整公式与正文间的段后距离

% paper title
% 缩写 DyLSKNet
% 翔哥拍板这个名字
% \title{Towards Very High-Resolution Guided Downscaling of Land Surface Temperature via Multi-Modal Images: A New Dataset, Method, and Toolkit}
% \title{MoCoLSK: Towards Very High-Resolution Guided Downscaling of Land Surface Temperature with Open-Source Benchmark and Toolkit}

% \title{GrokLST: Towards High-Resolution Benchmark and Toolkit for Land Surface Temperature Downscaling
% }

\title{
MoCoLSK: Modality Conditioned High-Resolution Downscaling for Land Surface Temperature
}
% Towards Very High-Resolution Guided Downscaling of Land Surface Temperature with Open-Source Benchmark and Toolkit

% \title{
% Towards Open-Source Benchmark  for High-Resolution Land Surface Temperature Downscaling
% }
% \title{Modality Conditioned Large Selective Kernel Network: Towards Very High-Resolution Guided Downscaling of Land Surface Temperature with Open-Source Benchmark and Toolkit}
% \title{Modality Conditioned LSKNet: Towards High-Resolution LST Downscaling with Open-Source Benchmark and Toolkit}

% and ToolKit 
% \title{Dynamic Guided Downscaling of Land Surface Temperature via Multi-Modal Imagery: A New Dataset, Method, and Toolkit}
% \title{Towards High-Resolution Guided Downscaling of Land Surface Temperature via SDGSAT-1 Imagery: A New Dataset, Method, and Toolkit}
% \title{Towards High-Resolution Guided Downscaling of Land Surface Temperature via SDGSAT-1 Imagery: A New Dataset, Method, and Toolkit}
% \title{Deep Guided Downscaling of Land Surface Temperature via Multi-Modal Imagery: A New Dataset, Method, and Toolkit}

% author names and IEEE memberships
% !TEX root = ../main.tex

% author names and IEEE memberships
\author{
  Qun~Dai,
  Chunyang~Yuan,
  Yimian~Dai,  
  Yuxuan~Li,
  Xiang~Li,
  Kang~Ni,
  Jianhui~Xu,
  Xiangbo Shu,
  Jian~Yang
  
  \thanks{
    This work was supported by the National Natural Science Foundation of China (62301261, % 我
       62206134, % 翔哥
       62101280, % 倪康
       62222207, 62332010, 62427808, % 舒老师
       U24A20330, 62361166670), % 杨老师
    % 博后面上
    China Postdoctoral Science Foundation (2021M701727, 2023M731781), % 倪康
    and the GDAS' Project of Science and Technology Development (2023GDASZH-2023010101).
    \emph{The first two authors contributed equally to this work. (Corresponding author:
      Yimian Dai, Kang Ni, Jianhui Xu)}
    }

  % 南理工
  \thanks{
    % Yimian Dai, Lingfeng Yang, and Jian Yang 
    Qun Dai and Xiangbo Shu are with School of Computer Science and Engineering, Nanjing University of Science and Technology, Nanjing, China.
    (e-mail:
    \href{mailto:qun.dai.grokcv@gmail.com}{qun.dai.grokcv@gmail.com};    
    % \href{mailto:yimian.dai@gmail.com}{yimian.dai@gmail.com};
    \href{mailto:shuxb@njust.edu.cn}{shuxb@njust.edu.cn}).
    % \href{mailto:csjyang@njust.edu.cn}{csjyang@mail.njust.edu.cn}
  }

% 南邮
\thanks{
    Chunyang Yuan and Kang Ni are with School of Computer Science and Technology, Nanjing University of Posts and Telecommunications, Nanjing, China. 
    % Kang Ni is also with Key Laboratory of Radar Imaging and Microwave Photonics, Nanjing University of Aeronautics and Astronautics, Ministry of Education, Nanjing, China.
    (e-mail:
    \href{mailto:chunyang.yuan.cs@gmail.com}{chunyang.yuan.cs@gmail.com};
    \href{mailto:tznikang@163.com}{tznikang@163.com}).
  }
  
  % % 南开
  \thanks{
  Yimian Dai, Yuxuan Li, Xiang Li, and Jian Yang are with PCA Lab, VCIP, College of Computer Science, Nankai University. 
  Xiang Li also holds a position at the NKIARI, Shenzhen Futian. 
  (e-mail:
  \href{mailto:yimian.dai@gmail.com}{yimian.dai@gmail.com};
  \href{mailto:yuxuan.li.17@ucl.ac.uk}{yuxuan.li.17@ucl.ac.uk};
  \href{mailto:xiang.li.implus@nankai.edu.cn}{xiang.li.implus@nankai.edu.cn};
  \href{mailto:csjyang@nankai.edu.cn}{csjyang@nankai.edu.cn})
  }

% 广州地理所
  \thanks{
  Jianhui Xu 
  % and Xia Zhou
  % , and Na Li 
  % are 
  % is with Key Lab of Guangdong for Utilization of Remote Sensing and Geographical Information System, Guangdong Open Laboratory of Geospatial Information Technology and Application, Guangdong Engineering Technology Research Center of Remote Sensing Big Data Application, Guangzhou Institute of Geography, Guangdong Academy of Sciences, Guangzhou 510070, China.
  is with Guangdong Provincial Key Laboratory of Utilization Remote Sensing and Geographical Information System, Guangzhou Institute of Geography, Guangdong Academy of Sciences, Guangzhou, 510070, China.
  % Na Li is also with State Key Laboratory of Organic Geochemistry, CAS Center for Excellence in Deep Earth Science, Guangzhou Institute of Geochemistry, Chinese Academy of Sciences, Guangzhou 510640, China.
    (e-mail:
    \href{mailto:xujianhui306@gdas.ac.cn}{xujianhui306@gdas.ac.cn}).
  }

}

\maketitle

% !TEX root = ../main.tex

\begin{abstract}
% Land Surface Temperature (LST) is a critical parameter for environmental studies, but obtaining high-resolution LST data remains challenging due to the spatio-temporal trade-off in satellite remote sensing.
% Guided LST downscaling has emerged as a solution, but current methods often neglect spatial non-stationarity and lack a open-source ecosystem for deep learning methods. 
% To address these limitations, we propose the Modality-Conditional Large Selective Kernel (MoCoLSK) Networks, a novel architecture that dynamically fuses multi-modal data through modality-conditioned projections. MoCoLSK re-engineers our previous LSKNet to achieve a confluence of dynamic receptive field adjustment and multi-modal feature integration, leading to enhanced LST prediction accuracy. 
% Furthermore, we establish the GrokLST project, a comprehensive open-source ecosystem featuring the GrokLST dataset, a high-resolution benchmark, and the GrokLST toolkit, an open-source PyTorch-based toolkit encapsulating MoCoLSK alongside 40+ state-of-the-art approaches.
% Extensive experimental results validate MoCoLSK's effectiveness in capturing complex dependencies and subtle variations within multispectral data, outperforming existing methods in LST downscaling.
% Our code, dataset, and toolkit are available at \url{https://github.com/GrokCV/GrokLST}.

% content in original manuscript，reviewer 2-comment 7，添加了“spatial”。
% Land Surface Temperature (LST) is a critical parameter for environmental studies, but directly obtaining high-resolution LST data remains challenging due to the spatio-temporal trade-off in satellite remote sensing.
Land Surface Temperature (LST) is a critical parameter for environmental studies, but directly obtaining high spatial resolution LST data remains challenging due to the spatio-temporal trade-off in satellite remote sensing.
Guided LST downscaling has emerged as an alternative solution to overcome these limitations, but current methods often neglect spatial non-stationarity, and there is a lack of an open-source ecosystem for deep learning methods.
In this paper, we propose the Modality-Conditional Large Selective Kernel (MoCoLSK) Network, a novel architecture that dynamically fuses multi-modal data through modality-conditioned projections. MoCoLSK achieves a confluence of dynamic receptive field adjustment and multi-modal feature fusion, leading to enhanced LST prediction accuracy. 
Furthermore, we establish the GrokLST project, a comprehensive open-source ecosystem featuring the GrokLST dataset, a high-resolution benchmark, and the GrokLST toolkit, an open-source PyTorch-based toolkit encapsulating MoCoLSK alongside 40+ state-of-the-art approaches.
Extensive experimental results validate MoCoLSK's effectiveness in capturing complex dependencies and subtle variations within multispectral data, outperforming existing methods in LST downscaling.
Our code, dataset, and toolkit are available at \url{https://github.com/GrokCV/GrokLST}.
\end{abstract}

\vspace{-1\baselineskip}
% !TEX root = ../main.tex
% \bibliography{../reference.bib}

\section{Introduction} \label{sec:introduction}

Land Surface Temperature (LST) reflects the complex mass and energy exchanges between the Earth's surface and the atmosphere \cite{NCC2023forest}. 
% \cite{RSE2022generating, RG2023LSTReview, NCC2023forest}. 
It serves as a critical indicator for evaluating ecological and climatic dynamics across various scales and plays a vital role in environmental studies, including urban heat island analysis \cite{RSE2021urban}, forest fire monitoring, land surface evapotranspiration \cite{RSE2022evapotranspiration}, soil moisture inversion \cite{RSE2024LSE}, and geothermal anomaly detection. However, the inherent limitations of satellite remote sensing hinder the acquisition of high spatial resolution LST data, specifically the unavoidable trade-off between temporal and spatial resolutions \cite{JAG2024duallayer}. For instance, Landsat 8 offers a spatial resolution of 100 meters but revisits the same region only once every 16 days \cite{TGRS2023TransferLearningLandsat}. In contrast, MODIS provides observations twice a day but at a coarser spatial resolution of 1 kilometer \cite{ESSD2024TRIMSLST}.
To address this challenge, one approach is to optimize sensing instruments and enhance satellite data transmission capabilities, but this is costly and time-consuming \cite{RSE2022nonlineardownscaling}. A more feasible alternative is to develop LST downscaling models.

Downscaling refers to transforming low-resolution (LR) images into high-resolution (HR) ones to enhance spatial detail information \cite{JAG2024duallayer}. Over the past two decades, various LST downscaling techniques have emerged, primarily categorized into statistical regression models, machine learning-based models, fusion models, and physical models \cite{JAG2024duallayer}. Classical linear statistical models, such as Disaggregation of Radiometric Surface Temperature (DisTrad) \cite{RSE2003DisTrad} and Thermal Sharpening (TsHARP) \cite{RSE2007TsHARP}, rely on the scale-invariant relationship between the Normalized Difference Vegetation Index (NDVI) and LST, employing global regression for downscaling. However, since LST is influenced by multiple factors such as wind, terrain, and land cover types, using a single biophysical parameter, like NDVI, as a predictor is insufficient \cite{RSE2022nonlineardownscaling}. To address this limitation, machine learning-based models, such as Random Forest (RF) \cite{JSTARS2019RF} and Extreme Gradient Boosting (XGBoost) \cite{RSE2021XGBoost}, leverage multiple biophysical parameters to effectively achieve LST downscaling while mitigating the risk of overfitting \cite{RSE2022nonlineardownscaling}. However, these models primarily adopt global regression paradigms, which perform well in homogeneous areas but often fall short in highly heterogeneous regions, such as urban environments \cite{RSE2022nonlineardownscaling}. Methods like Geographically Weighted Regression (GWR) \cite{TGRS2016GWR}  
% , Multi-Factor Geographically Weighted Regression (MFGWR) \cite{JSTARS2019MFGWR},
and Multiscale Geographically Weighted Regression (MGWR) \cite{AAAG2017MGWR} effectively address the spatial heterogeneity of LST. Additionally, Geographically and Temporally Weighted Regression (GTWR) \cite{TGRS2019Regression} models the spatiotemporal non-stationarity between LST and environmental factors in time-series datasets. On the physical modeling front, the DTsEB method \cite{RSE2023physical}, based on the Surface Energy Balance (SEB), explains the interactions between biophysical parameters and LST from a physical mechanism perspective.

Recently, deep learning has catalyzed a paradigm shift in computer vision and remote sensing, also markedly affecting LST downscaling \cite{ECCV2014SRCNN, lim2017enhanced, CVPR2018RDN, yuan2023recurrent, wang2024sgnet}.
These models leverage the ability of deep neural networks to learn complex spatial and temporal patterns from data, enabling them to effectively capture the relationships between LR and HR data.
% ==========================原稿内容 start
% For instance, Wang and Tian pioneered the Super Resolution Deep Residual Network (SRDRN) to refine the downscaling of daily meteorological parameters \cite{wang2022deep}, vastly outstripping conventional methods. 
% Concurrently, Mukherjee and Liu advanced an encoder-decoder super-resolution framework \cite{mukherjee2021downscaling}, incorporating self-attention to increase the resolution of MODIS spectral bands with notable efficacy.
% ===============================原稿内容 end

%  ---------------------校正版
% 深度学习用于降尺度
In the context of HR climate data generation, Wang \textit{et al.}\ pioneered the Super Resolution Deep Residual Network (SRDRN) to refine the downscaling of daily meteorological parameters like precipitation and temperature \cite{WR2021SRDRN}, vastly outstripping conventional methods. 
Building upon this, Mital \textit{et al.}\ contributed a fine-scale (400 m) dataset, achieved via a data-driven downscaling model that discerned the impact of topography on climate variables \cite{ESSD2022Precipitation}. 
Furthermore, Vaughan \textit{et al.}\ introduced convolutional conditional neural processes, a versatile deep learning framework for multisite statistical downscaling, which enabled the generation of continuous stochastic forecasts for climate variables across any geographic location \cite{GDM2022Convolutional}. 
% Shifting focus to remote sensing data enhancement, 
% 空时融合，reviewer#1-comment-1
Unlike methods that focus solely on spatial downscaling, Geographically and Temporally Neural Network Weighted Autoregression (GTNNWAR) \cite{ISPRS2022autoregressive} uses a two-stage deep neural network and autoregressive model to downscale MODIS LST from 1 km to 100 m through spatiotemporal fusion.
Yu \textit{et al.}\ introduced a DisTrad-Super-Resolution Convolutional Network, integrating statistical methods with deep learning to significantly improve the spatial and temporal resolutions of remote sensing imagery, enabling more refined analysis of lake surface temperature dynamics \cite{JSTARS2021TemperatureDownscaling}.
% This advancement allowed for a more refined analysis of lake surface temperature dynamics.
Compared to methods that only consider spatial non-stationarity, they additionally leverage temporal information from time-series data and model its temporal non-stationarity, offering a greater advantage in LST downscaling.
% Transformer 降尺度方法，reviewer#1-comment-2
Besides, Mukherjee and Liu developed an encoder-decoder super-resolution architecture that incorporated a custom loss function and a self-attention mechanism, adeptly increasing the resolution of MODIS spectral bands while maintaining spatial and spectral fidelity without supplementary spatial inputs \cite{mukherjee2021downscaling}.
Although significant progress has been made in downscaling meteorological data using deep learning methods, there are still challenges, such as the lack of effective dynamic fusion architectures and a specialized open-source ecosystem for downscaling.

Shifting focus to the field of computer vision, super-resolution (SR) aligns closely with the concept of downscaling. SR can be categorized into single-image SR (SISR)~\cite{ECCV2014SRCNN,lim2017enhanced,CVPR2018RDN,zhang2018image,ICCVW2021SwinIR,ICCV2023SRFormer,ECCV2025HiT-SR} and guided image SR (GISR)~\cite{zhao2022discrete,metzger2023guided,yuan2023recurrent,xiang2022detail,zhou2023memory}. GISR aims to restore HR images from LR ones by leveraging structural information from HR guidance images of the same scene, while SISR does not rely on these HR guidance data. 
% SISR
Super-Resolution Convolutional Neural Network (SRCNN) ~\cite{ECCV2014SRCNN}, the first SISR method to learn the mapping between high-resolution and low-resolution images in an end-to-end manner, significantly boosted the development of the SR field. 
% The Enhanced Deep Super-Resolution Network (EDSR)~\cite{lim2017enhanced} optimized unnecessary residual modules and demonstrated superior reconstruction performance by winning the NTIRE2017 Super-Resolution Challenge~\cite{CVPRW2017NTIRE}. 
% To fully exploit the hierarchical features of low-resolution images, RDN adopts a dense residual strategy, achieving favorable performance. 
To fully exploit the hierarchical features of LR images, Residual Dense Network (RDN)~\cite{CVPR2018RDN} adopts a dense residual strategy, effectively leveraging the hierarchical information from the original LR images, achieving excellent performance. 
Additionally, Residual Channel Attention Networks (RCAN)~\cite{zhang2018image} introduces channel attention mechanisms into deep residual networks, resulting in improved accuracy and enhanced visual quality. 
% Transformer SISR
In recent years, many works have incorporated Transformer into SR tasks, such as SwinIR~\cite{ICCVW2021SwinIR}, DAT~\cite{ICCV2023DAT}, and SRFormer~\cite{ICCV2023SRFormer}, benefiting from the global receptive field of the self-attention mechanism. However, the quadratic complexity and the need for large-scale training data remain challenges. Zhang \textit{et al.}\ \cite{ECCV2025HiT-SR} proposed a general strategy to convert Transformer-based SR networks into Hierarchical Transformer (HiT-SR), enhancing SR performance through multi-scale features while maintaining an efficient design.

% GISR
Recent advances in GISR can be broadly categorized into two main approaches: cross-modal feature fusion and shared-private feature separation. 
Cross-modal feature fusion methods focus on effectively combining information from the target and guidance images. 
For instance, Zhong \textit{et al.}\ \cite{zhong2021high} introduced an attention-based hierarchical multi-modal fusion strategy that selected structurally consistent features. 
Building upon this, Shi \textit{et al.} \cite{shi2022symmetric} proposed a symmetric uncertainty-aware transformation to filter out harmful information from the guidance image, ensuring more reliable feature fusion. 
Furthermore, Wang \textit{et al.} \cite{wang2024sgnet} developed a structure-guided method that propagates high-frequency components from the guidance to the target image in both the frequency and gradient domains, enabling more comprehensive fusion of structural details. 
On the other hand, shared-private feature separation methods aim to disentangle the common and unique information between the target and guidance images. 
Deng \textit{et al.} \cite{deng2020deep} employed convolutional sparse coding to split the shared and private information across different modalities, facilitating more targeted feature fusion. 
Building on this concept, He \textit{et al.} \cite{he2021towards} separated RGB features into high-frequency and low-frequency components using octave convolution, allowing for more fine-grained information integration. 
% Similarly, Zhu \textit{et al.} \cite{zhu2023probability} proposed a probabilistic global cross-modal upsampling approach to learn the cross-modal information of the guidance image, enabling more robust feature fusion. 
Additionally, Xiang \textit{et al.} \cite{xiang2022detail} introduced a detail injection fusion network to fully utilize the nonlinear complementary features of both the target and guidance images, achieving more effective detail restoration.
Existing GISR methods typically rely on fixed receptive fields and vanilla multimodal fusion methods (e.g., addition, concatenation), which may fail to effectively capture the multi-scale dependencies in LST data and the complex interactions between different modalities.
Despite these advancements, LST downscaling has not kept pace with the rapid developments seen in SISR and GISR \cite{ISPRS2024systematicreview}. This stagnation may stem from a lack of a supportive ecosystem for deep learning innovation, with two primary obstacles identified:
\begin{enumerate}
    \item \textbf{Absence of High-Resolution Benchmark Dataset:} 
    Satellite data disparities in region, time, and sensor selection hinder methodological comparisons. The lack of uniformity in satellite data selection and the scarcity of \textit{ HR thermal infrared data} ($\le 30~\mathrm{m}$) pose significant challenges for LST SR research. 
    Therefore, establishing a standardized HR benchmark dataset are crucial for advancing the field. 
    \item \textbf{Scarcity of Open-Source LST SR Toolkit:} The absence of a dedicated open-source toolkit for LST SR hinders the community's ability to replicate, refine, and challenge existing methods. Such a toolkit would be essential for fostering collaborative development and accelerating progress in the field.
\end{enumerate}
  % \item[1)] \textbf{The absence of a standardized benchmark dataset}: While portions of satellite data are publicly accessible, the inconsistency in the selection of regions, time frames, and sensors across different studies has led to an incommensurability of results, hampering the objective comparison of methods within the literature.
%   \item[2)] \textbf{The scarcity of an open-source code repository}: A communal codebase serves as an invaluable resource for researchers striving to replicate and benchmark diverse methods. However, existing repositories cater predominantly to single-frame natural image SR, leaving LST SR tasks without a dedicated platform for methodological exchange and evaluation.
% \end{itemize}

Moreover, most current deep learning models for LST downscaling are straightforward adaptations from GISR models in computer vision, without fully considering the unique characteristics of LST data and its associated challenges \cite{ISPRS2024systematicreview}. 
\textit{As the resolution of thermal infrared bands reaches high levels ($\le 30~\mathrm{m}$)}, small-scale local features, such as buildings and roads, emerge alongside large-scale land cover types like water bodies, deserts, and grasslands.
These local features are prone to mixing with their surroundings, introducing additional complexity to the downscaling process.
According to an analysis of the HR LST data, we identify two primary limitations in existing methodologies:
\begin{enumerate}
    \item \textbf{Inability to Dynamically Adjust Receptive Fields:} 
    The stark spatial heterogeneity of LST necessitates a model capable of adjusting its receptive field to the diverse scales of temperature fluctuations. This adaptability is crucial for accurately capturing the local contrasts within LST distributions over various spatial extents.
    \item \textbf{Multi-modal Fusion in a Uni-dimensional Manner:} 
    Existing approaches to integrating multi-modal auxiliary data with LST features have been restricted to simplistic, uni-dimensional operations, such as addition, multiplication, or concatenation. These approaches do not suffice to unravel the complex interdependencies within HR guidance data.
    % ,leading to suboptimal feature enhancement and an underutilization of the rich information available in multi-modal sources.
\end{enumerate} 
To address these challenges, we propose the \textbf{Modal-Conditioned Large Selective Kernel (MoCoLSK) Network}, a novel dynamic multimodal fusion framework.
MoCoLSK builds upon our previous Large Selective Kernel Network (LSKNet)~\cite{ICCV2023LSK} by replacing the static convolution in the kernel selection mechanism with a dynamic modal-conditioned projection. This projection is determined jointly by coarse-resolution LST and fine-resolution guidance data, enabling dynamic receptive field adjustment. Consequently, MoCoLSK adaptively learns fine-grained, discriminative texture features, precisely modeling the mapping between coarse-resolution LST and fine-resolution guidance data, thereby enhancing the accuracy of LST downscaling.
% ---------------------------------------矫正版

Furthermore, to foster research and advancement in LST downscaling, we establish \textbf{a comprehensive open-source ecosystem termed the GrokLST project}.
Our contributions include the \textbf{GrokLST dataset}, a benchmark featuring 641 pairs of LR and HR LST images from the SDGSAT-1 satellite data, along with corresponding auxiliary data of multiple modalities.
Accompanying the dataset is \textbf{GrokLST toolkit}, an open-source PyTorch-based toolkit encapsulating our MoCoLSK model alongside other \textbf{40+} state-of-the-art approaches, empowering researchers to effortlessly leverage the GrokLST dataset and conduct standardized evaluations.

Through extensive experimental results, we validate the effectiveness of MoCoLSK, showcasing its ability to capture the complex dependencies and subtle variations within multispectral data, outperforming existing methods in LST downscaling. The proposed MoCoLSK architecture and the GrokLST ecosystem pave the way for advancing research and applications in HR LST retrieval, providing a solid foundation for future developments in this domain.

\section{GrokLST: Open-Source Ecosystem} 
\label{sec:ecosystem}

% \subsection{HeiheLST Dataset} 
\subsection{GrokLST Dataset} 
\label{subsec:dataset}

\begin{figure*}[htbp]
  \centering
    \vspace{-1\baselineskip}
  \includegraphics[width=.95\textwidth]{"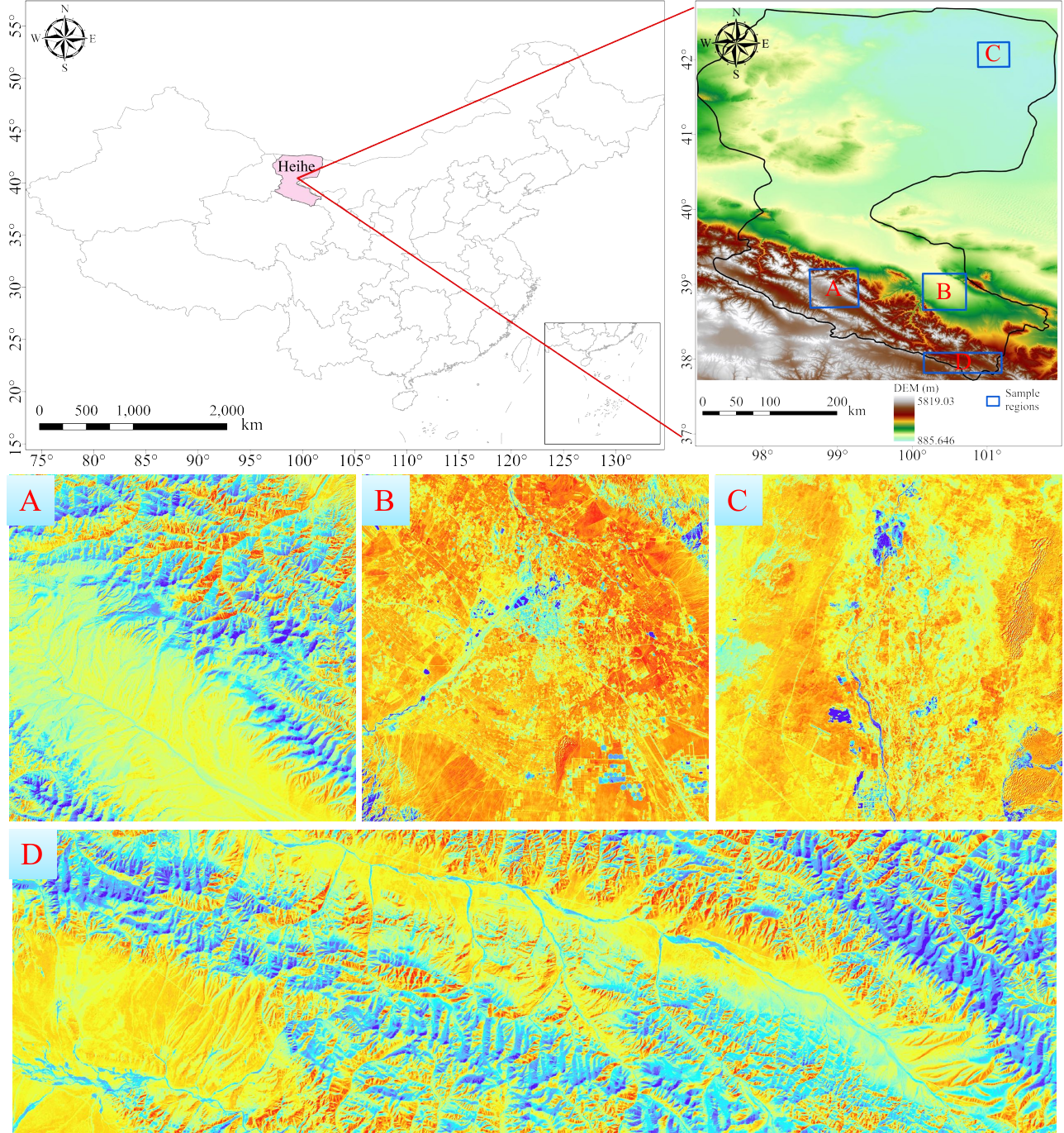"}
      % \vspace{-.5\baselineskip}
  \caption{Selection area and representative images of our GrokLST dataset. As demonstrated, upon reaching a resolution of 30 meters or lower, numerous local details emerge. This phenomenon underscores the critical need for models with a dynamic receptive field capable of capturing these intricate patterns.}
  \label{fig:study-area}
  \vspace{-1\baselineskip}
\end{figure*}

% Recent proliferation of accessible satellite imagery has catalyzed the development of deep learning models in the domain of thermal remote sensing. 
% However, the disparate preprocessing practices and dataset structures across different research efforts impedes the uniform evaluation of emerging models.
% Recognizing the need for consistency in model evaluation, we create \textit{GrokLST}, an open-source benchmark dataset tailored for LST downscaling, aiming to serve as a cornerstone for the field's advancement through standardized algorithm comparisons.

The recent proliferation of accessible satellite imagery has catalyzed the development of deep learning models in the domain of thermal remote sensing. However, the field of LST downscaling currently lacks HR open-source datasets, which hinders the comprehensive evaluation and comparison of emerging models. Moreover, the disparate preprocessing practices and dataset structures across different research efforts further impede the uniform assessment of state-of-the-art techniques. 

Recognizing the need for consistency in model evaluation and the importance of HR data, we introduce \textit{GrokLST}, an open-source benchmark dataset specifically designed for LST downscaling. GrokLST fills a critical gap in the field by providing a HR dataset that enables researchers to evaluate and compare their models on a standardized platform, fostering the advancement of LST downscaling through rigorous and consistent algorithm assessments.

\subsubsection{Study Area}

% Fig.~\ref{fig:study-area} illustrates the spatial distribution of the Heihe River Basin, the second-largest inland river basin in Northwestern China, serving as the focal point of this study. 
% Encompassed within the geographical coordinates of 98° to 101°E longitude and 38° to 42°N latitude, the basin lies at the heart of the Hexi Corridor and is the predominant inland watershed in Western Gansu and Qinghai provinces. 

As depicted in Fig.~\ref{fig:study-area}, the pivotal focus of this study is the Heihe River Basin, the second-largest inland river basin in Northwestern China. Geographically positioned between 98° to 101°E longitude and 38° to 42°N latitude, the basin is nestled within the Hexi Corridor, serving as the primary inland watershed in Western Gansu and Qinghai provinces. 

The Heihe River Basin's unique positioning amidst the Eurasian landmass and its adjacency to towering mountain ranges bestow upon it a distinct continental climate. This climate is predominantly shaped by the mid-to-high latitude westerly wind circulation and periodic influxes of polar cold air masses. The basin is characterized by its arid conditions, punctuated by sparse and concentrated precipitation, frequent high winds, abundant sunshine, intense solar radiation, and significant diurnal temperature variations. Spanning 821 kilometers from its source to its terminus at Lake Juyan, the Heihe River carves its path through three distinct ecological environments, covering an area of approximately 142,900 square kilometers. 

The intricate interplay of climatic factors and geographical diversity renders the Heihe River Basin a prime candidate for environmental remote sensing and land surface temperature downscaling studies. Its vast and varied land covers, which include impervious urban structures, verdant vegetation, and sprawling water bodies, provide a diverse palette for implementing advanced deep learning techniques and computational vision approaches. These methods are employed to super-resolve imagery, facilitating a granular environmental analysis, and thereby highlighting the unique value of this study area.

\subsubsection{Data Source and Preparation}

Our GrokLST dataset leverages the cutting-edge remote sensing capabilities of the Sustainable Development Goals Science Satellite 1 (SDGSAT-1), which was launched on November 5, 2021, to bolster the United Nations Sustainable Development Goals \cite{SCIB2023SDGSAT}.
SDGSAT-1's Multispectral Imager for Inshore (MII) and Thermal Infrared Spectrometer (TIS) sensors synergistically contribute to this dataset, with their spectral characteristics and band designations detailed in Tab.~\ref{tab:SDGResolution}.

\begin{table}[htbp]
  \renewcommand\arraystretch{1.2}
  \centering
  \caption{SDGSAT-1 multispectral and TIR bands used in creation of our GrokLST dataset.}
  \label{tab:SDGResolution}
  \setlength{\tabcolsep}{4.pt}
\begin{tabular}{@{}ccccl@{}}
\toprule
\multirow{2}{*}{Sensor} & Band & Bandwidth & Resolution & \multirow{2}{*}{Note} \\
& Name & ($\mu$m) & (m) & \\
\midrule
% Band Name & Bandwidth ($\mu$m) & Resolution (m) & Note \\ 
% \multirow{7}{*}{MII} & Band 1    & 0.374 $\sim$ 0.427    & 10             \\
\multirow{7}{*}{MII} & Band 2    & 0.410 $\sim$ 0.467    & 10 & Deep Blue            \\
                     & Band 3    & 0.457 $\sim$ 0.529    & 10 & Blue            \\
                     & Band 4    & 0.510 $\sim$ 0.597    & 10 & Green            \\
                     & Band 5    & 0.618 $\sim$ 0.696    & 10 & Red            \\
                     & Band 6    & 0.744 $\sim$ 0.813    & 10 & VRE            \\
                     & Band 7    & 0.798 $\sim$ 0.911    & 10 & NIR            \\ \midrule
\multirow{3}{*}{TIS} & Band 1    & 8.0 $\sim$ 10.5       & 30 &             \\
                     & Band 2    & 10.3 $\sim$ 11.3      & 30 &             \\
                     & Band 3    & 11.5 $\sim$ 12.5      & 30 &             \\ \bottomrule
\end{tabular}
\vspace{-1\baselineskip}
\end{table}

For the specific LST retrieval algorithm of SDGSAT-1, please refer to our latest work \cite{TGRS2024Retrieval}. 
The validation of the LST retrieval accuracy against in-situ measurements from the HiWATER sites, available at the National Cryosphere Desert Data Center (\url{http://www.ncdc.ac.cn}), reveals an RMSE of 2.598 K and an $R^2$ of 0.977. 
% This dataset augments the existing LST downscaling resources and underpins HR environmental monitoring in alignment with the Sustainable Development Goals.

% Our LST downscaling dataset incorporates a suite of ten spectral bands and indices from SDGSAT-1, enhancing the algorithm's precision. The selected bands include ``B2 Deepblue'', ``B3 Blue'', ``B4 Green'', ``B5 Red'', and ``B7 NIR'', while key indices feature the digital elevation model (DEM), NDWI, NDVI, and Normalized Difference Moisture Vegetation Index (NDMVI). These elements are pivotal for capturing detailed environmental characteristics, such as vegetation vigor and water content. 
% The combination of LST data and these auxiliary inputs forms a comprehensive dataset that significantly enhances the SDGSAT-based LST downscaling algorithm.

% Building upon these advancements, our GrokLST dataset incorporates a comprehensive suite of ten spectral bands and indices from SDGSAT-1, enhancing the algorithm's precision. The selected bands include ``B2 Deepblue'', ``B3 Blue'', ``B4 Green'', ``B5 Red'', and ``B7 NIR'', while key indices feature the digital elevation model (DEM), NDWI, NDVI, and Normalized Difference Moisture Vegetation Index (NDMVI). These elements are pivotal for capturing detailed environmental characteristics, such as vegetation vigor and water content. The combination of LST data and these auxiliary inputs forms a rich dataset that significantly enhances the SDGSAT-based LST downscaling algorithm, setting it apart from previous approaches that relied on a more limited set of input features.

\subsubsection{Dataset Description}

The GrokLST dataset includes 10 different types of HR (30m) auxiliary data. The selected bands include ``B2 Deepblue'', ``B3 Blue'', ``B4 Green'', ``B5 Red'', ``B6 VRE'', and ``B7 NIR'', while key indices feature the Digital Elevation Model (DEM), Normalized Difference Water Index (NDWI), NDVI, and the Normalized Difference Moisture Vegetation Index (NDMVI). These auxiliary data play a critical role in enriching the contextual understanding required for accurate LST modeling. Moreover, the LST data is provided at a resolution of 30 meters, offering detailed thermal spectral profiles, making it suitable for high-precision studies. 
Specifically, the GrokLST dataset consists of 641 pairs of image data from the Heihe River Basin, covering four different scales (i.e., 30m, 60m, 120m, and 240m), including both LST data and HR guidance data, to address downscaling challenges across various scales. The detailed dataset dimensions and experimental setup can be found in \ref{subsubsec:V.A.dataset}.

\subsection{GrokLST Toolkit} \label{subsec:toolkit}

The field of LST downscaling has long been hindered by a lack of accessible, open-source tools that foster innovation and reproducibility.
To address this gap, we introduce GrokLST, a comprehensive deep learning toolkit designed specifically for LST downscaling tasks. Built on the PyTorch framework, GrokLST offers high flexibility and speed in model development and training, drawing inspiration from proven architectures in generic computer vision toolboxes like MMDetection and Detectron2.

% GrokLST is built on the PyTorch framework, allowing for high flexibility and speed in model development and training. The toolkit draws inspiration from proven architectures in generic computer vision toolboxes like MMDetection and Detectron2 but is uniquely tailored to meet the specific demands of LST downscaling. GrokLST distinguishes itself through several key features:
GrokLST distinguishes itself through several key features that cater to the unique demands of LST downscaling.
\begin{enumerate}
    \item \textbf{Comprehensive Model Support:} GrokLST provides out-of-the-box support for over 40 state-of-the-art super-resolution models. This extensive library not only facilitates easy comparison of different methods but also serves as a foundation for further research and development.
    \item \textbf{Customizable Components:} Unlike general-purpose toolkits, GrokLST offers enhanced flexibility in model configuration. Users can choose from a variety of backbones, necks, and attention mechanisms, tailoring the architecture to specific LST downscaling needs.
    \item \textbf{Specialized Tools and Metrics:} The toolkit includes specialized dataset loaders, data augmentation pipelines, and LST-specific evaluation metrics. These components are essential for accurately assessing model performance under diverse environmental conditions.
\end{enumerate}
% In summary, GrokLST is a powerful and user-friendly toolkit that addresses the long-standing need for accessible, open-source tools in the LST downscaling community.

% To advance the field's methodological standards, GrokLST emphasizes reproducibility. It provides pre-trained models, detailed training scripts, and comprehensive logs, enabling researchers to replicate and extend existing experiments. Additionally, a benchmarking study of various downscaling methods offers insights into their performance and optimal configurations.

% !TEX root = ../main.tex
% \bibliography{../reference.bib}

\section{Method} \label{sec:method}

\begin{figure*}[ht]
    \centering
    \vspace{-1\baselineskip}
    \includegraphics[width=.98\textwidth]{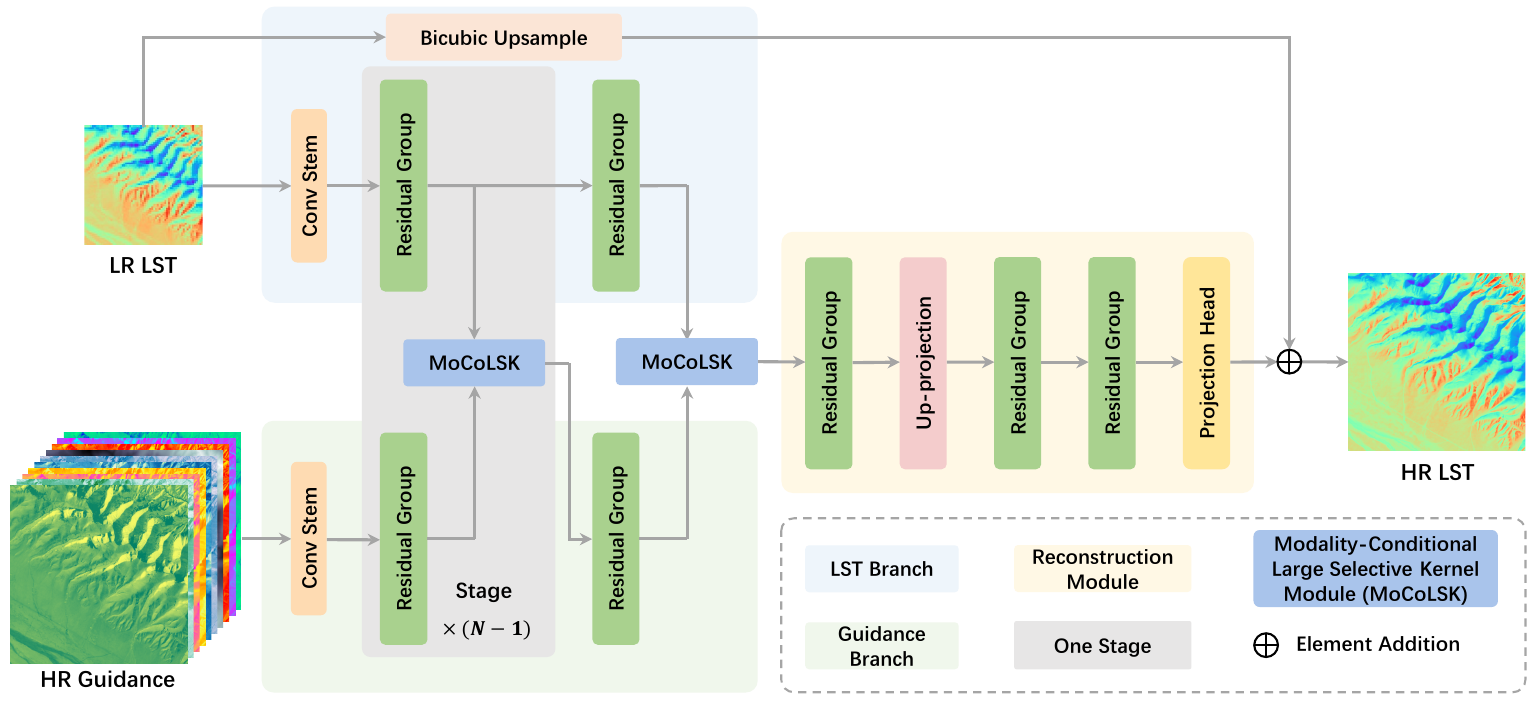}
    \caption{The overall framework of MoCoLSK-Net primarily includes LST branch, guidance branch, MoCoLSK module, and reconstruction module. MoCoLSK-Net stacks $N$ stages, with each stage comprising two residual groups and one MoCoLSK module. The output from the $N$-th stage is fed into the reconstruction module. The downscaled HR LST is obtained by adding output of the reconstruction module to bicubic upsampling result of the original LR LST data.}\label{fig:mocolsk-net}
    % \vspace{-1\baselineskip}
\end{figure*}

The problem of guided LST downscaling can be formulated as follows: Given a LR LST map $T_{lr} \in \mathbb{R}^{1 \times H \times W}$ and HR guided data $G_{hr} \in \mathbb{R}^{K \times sH \times sW}$, the goal is to estimate an HR LST map $T_{sr} \in \mathbb{R}^{1 \times sH \times sW}$ that approximates the true HR LST map $T_{hr} \in \mathbb{R}^{1 \times sH \times sW}$. Here, $H$ and $W$ denote the height and width of the LR LST map, $s$ is the scaling factor, and $K$ represents the number of channels in the guided data.

In recent years, deep learning has emerged as a powerful tool for LST downscaling \cite{ISPRS2018Sentinel2SR}. These methods leverage the ability of deep neural networks to learn intricate feature representations and model complex relationships between input data and the desired output. A typical deep learning-based LST downscaling model can be expressed as:
\begin{equation}
T_{sr} = \mathcal{F}(T_{lr}, G_{hr}; \theta),
\end{equation}
where $\mathcal{F}$ represents the deep neural network with learnable parameters $\theta$. The network takes the LR LST map $T_{lr}$ and the HR guided data $G_{hr}$ as inputs and generates the HR LST map $T_{sr}$. The network is trained on a dataset of paired LR-HR LST maps and HR guided data, with the objective of minimizing a $L_1$ loss function that measures the discrepancy between the predicted HR LST map and the ground truth, defined as:
\begin{equation}
L(\theta) = \frac{1}{N}\sum_{i=1}^{N} \left \Vert \mathcal{F}(T^i_{lr}, G^i_{hr}) - T^i_{hr} \right \Vert_1.
\end{equation}

\subsection{MoCoLSK-Net Architecture}

As depicted in Fig.~\ref{fig:mocolsk-net}, our proposed MoCoLSK-Net comprises four primary components: LST branch, guidance branch,  MoCoLSK module, and reconstruction module, each designed to process and refine environmental data effectively.

\textbf{LST and Guidance Branches}: 
Apart from the different inputs, these two branches are almost completely homogeneous in structure.
Each branch initiates with a convolutional stem that is responsible for extracting initial features from the input LR LST map or HR guidance image. 
Following the convolutional stem are $N$ stages of Residual Groups, where each stage consists of multiple residual blocks \cite{CVPR16ResNetV1} with channel attention \cite{ECCV18CBAM}. 
Additionally, the two branches differ in one aspect: the LR LST map in the LST branch is processed through a bicubic upsample layer to match the desired output resolution, which serves as the preliminary step for further refinement.

\textbf{MoCoLSK Module}:
The MoCoLSK module is the core component of our network, designed to perform dynamic multi-modal fusion. 
It takes as input the features from the corresponding stages of the LST and guidance branches, and performs dynamic multimodal fusion and refinement.
Please refer to Section \ref{subsec:mocolsk-module} fore more details.

\textbf{Reconstruction Module}:
The reconstruction module is responsible for aggregating the refined features from the MoCoLSK modules and generating final downscaled LST.
Among a series of residual groups, this stage employs a up-projection unit \cite{CVPR2018DBPN} to generate HR features. Finally, projection head, consisting of two convolutional layers and a LeakyReLU activation layer, works together with the bicubic interpolation results to generate the final downscaled LST.
% The final output is a high-fidelity HR LST image that provides enhanced spatial details crucial for accurate environmental assessment and decision-making.

\begin{figure*}[ht]
    \centering
    \vspace{-1\baselineskip}
    \includegraphics[width=.98\textwidth]{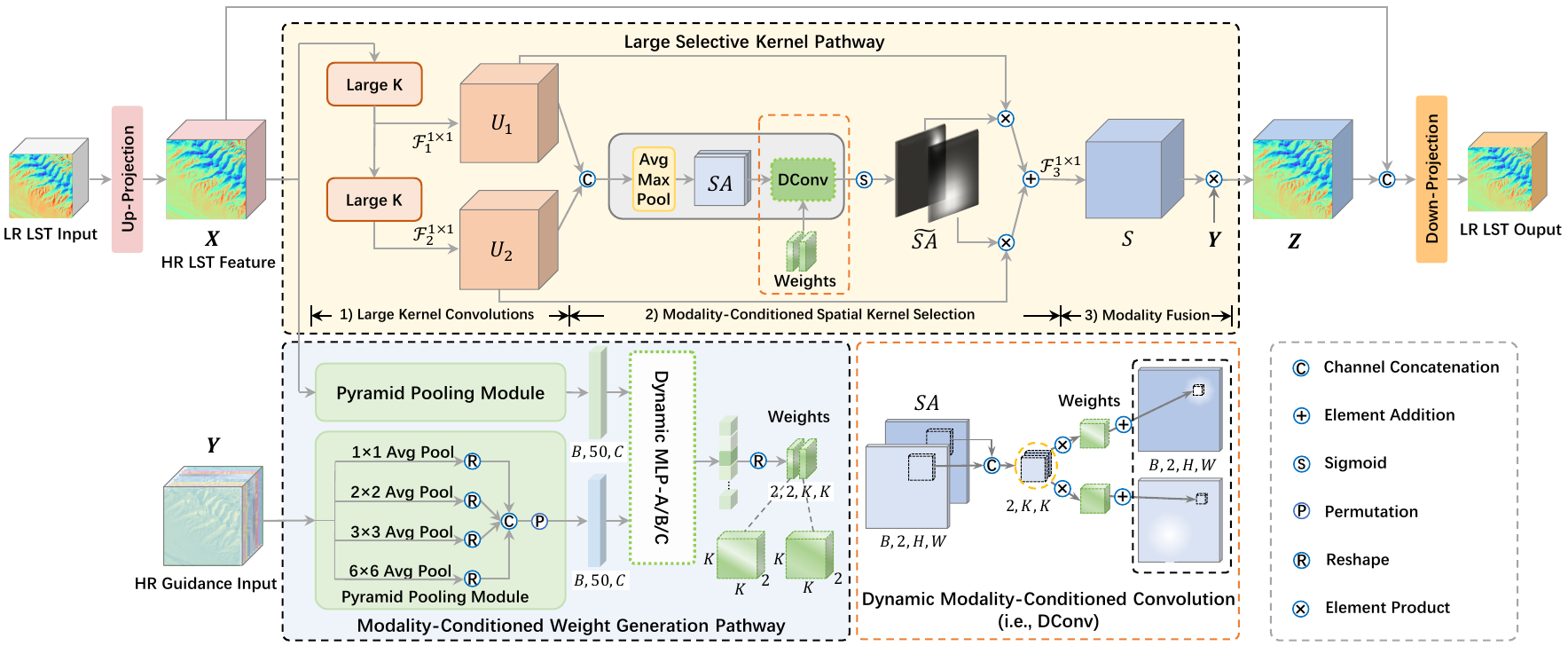}
    \caption{Overview of our proposed MoCoLSK Module. The MoCoLSK module primarily consists of the large selective kernel pathway and the modality-conditioned weight generation pathway. For the LST pathway, it essentially follows the original LSK \cite{ICCV2023LSK} module configuration but with two key differences: 1) the generation of the spatial selection mask $\bm{\widetilde{SA}}$ is modulated by the modality-conditioned weights from the MCWG pathway; 2) the output feature $\bm{Z}$ is the result of fusing two modality features. For the MCWG pathway, the HR LST features and HR guidance features are deeply fused using the pyramid pooling module and the dynamic MLP \cite{CVPR2022DMLP} to generate modality-conditioned weights. Additionally, to facilitate modality fusion and the stacking of multiple MoCoLSK modules for more refined LST feature reconstruction, we utilize up-projection and down-projection units to upsample the LR LST features and downsample the concatenated result of the modality fusion output $\bm{Z}$ with the HR LST features $\bm{X}$.}\label{fig:mocolsk-ss}
    \vspace{-1\baselineskip}
\end{figure*}

\subsection{MoCoLSK Module} \label{subsec:mocolsk-module}

As illustrated in Fig.~\ref{fig:mocolsk-ss}, the MoCoLSK module consists of two primary pathways: the Large Selective Kernel (LSK) pathway and the Modality-Conditioned Weight Generation (MCWG) pathway. 
Additionally, up and down projection layers \cite{CVPR2018DBPN} are introduced to perform upsampling and downsampling of LST features, enabling the framework to stack MoCoLSK multiple times for finer LST feature reconstruction.
The complete MoCoLSK module can be formulated as:
\begin{equation}
    \begin{aligned}
        T_{lr}^{(l)}  & = \text{MoCoLSK}(T_{lr}^{(l-1)}, G_{hr}^{(l-1)}), \\
                & = \text{Down} (\left [ \text{Up}(T_{lr}^{(l-1)}), \bm{Z} \right ]), 
    \end{aligned}
\end{equation}
where Up and Down refer to up-projection and down-projection layers \cite{CVPR2018DBPN}, respectively. 
$\bm{Z}$ is output feature of LSK pathway, $[\cdot]$ indicates channel concatenation, and $T_{lr}^{(l)}$ represents output of MoCoLSK at $l$-th stage.

Our MoCoLSK is based on LSK \cite{ICCV2023LSK} and aims to dynamicize the key static convolutions, achieving multimodal feature fusion through a dynamic receptive field driven by modality-conditioned weights. These weights are determined by dynamically fusing LST and guidance features through MCWG pathway and serve as convolution kernels for the dynamic modality-conditioned convolution (denoted as DConv), facilitating dynamic adjustments to the receptive field.

\subsubsection{Large Selective Kernel Pathway}\label{subsubsec:LSK-pathway}
LSK pathway closely follows the original LSK, with two key distinctions: first, the static convolution used to generate the spatial selection masks is replaced by DConv; second, the feature $\bm{S}$ is multiplied by HR guidance feature $\bm{Y}$ instead of $\bm{X}$.

Specifically, the LSK pathway takes HR LST features $\bm{X}$ and HR guidance features $\bm{Y}$ as inputs, and outputs the refined HR LST $\bm{Z}$ through three steps: (1) large kernel decomposition; (2) modality-conditioned spatial kernel selection; and (3) modality fusion.

\textbf{Large Kernel Decomposition:} 
The large kernel decomposition leverage HR LST feature $\bm{X}$ to generate large kernel features $U_1$ and $U_2$ at different scales, defined as follows:
\begin{equation}
    \begin{aligned}
        & \bm{U}_{1}=\mathcal {F}_{1}^{1\times1}(\mathcal {F}_{lk}^{5\times5}(\bm{X})), \\
        & \bm{U}_{2}=\mathcal {F}_{2}^{1\times1}(\mathcal {F}_{lk}^{7\times7}(\mathcal {F}_{lk}^{5\times5}(\bm{X}))),
    \end{aligned}
\end{equation}
where 
% $\bm{U}_{1}$ and $\bm{U}_{2}$ represent LST features with different large receptive fields. 
$\left \{ \mathcal {F}_{i}^{1\times1}, i=1,2. \right \} $ are point-wise convolutions. $\mathcal {F}_{lk}^{5\times5}$ and $\mathcal {F}_{lk}^{7\times7}$ denote depth-wise convolutions with kernel size 5, dilation 1 and kernel size 7, dilation 3, respectively.

\textbf{Modality-Conditioned Spatial Kernel Selection:} This selection aims to dynamically select features from spatial kernels with different receptive fields (i.e., $\bm{U}_{1}$ and $\bm{U}_{2}$) that are effective for refining HR LST feature, assisted by modality-conditioned weights generated through MCWG pathway.
Specifically, $\bm{U}_{1}$ and $\bm{U}_{2}$ are first concatenated along the channel dimension, followed by channel-wise average pooling $\mathcal P_{avg}$ and maximum pooling $\mathcal P_{max}$, and then concatenated again to obtain preliminary spatial attention weights $\bm{SA}$. This process is formulated as:
\begin{equation}
    \bm{SA}= \left [ \mathcal P_{avg}(\left [\bm{U}_{1},\bm{U}_{2}\right ]), \mathcal P_{max}(\left [\bm{U}_{1},\bm{U}_{2} \right ]) \right ].
\end{equation}

To obtain modality-conditioned spatial selection masks $\bm{\widetilde{SA}}$, we introduce a dynamic modality-conditioned convolution layer (denoted as $\bm{\mathcal F_{dconv}}$) powered by modality-conditioned weights from the MCWG pathway (see \ref{subsubsec:MCWG-pathway}), as detailed in the following:
\begin{equation}
    \bm{\widetilde{SA}}= \sigma(\bm{\mathcal F_{dconv}}^{2 \to {2} } (\bm{SA}, weights)),
    \label{equ:5}
\end{equation}
where superscript $(\cdot)^{2 \to 2}$ indicates that the number of channels remains 2. $\sigma$ is sigmoid activation function.

The large kernel features $\bm{U}_{1}$ and $\bm{U}_{2}$ are spatially weighted by their corresponding spatial masks (i.e., $\bm{\widetilde{SA}}_{1}$ and $\bm{\widetilde{SA}}_{2}$), then added and passed through a point-wise convolutional layer $\mathcal {F}_{3}^{1\times1}$ to obtain the dynamically selected features $\bm{S}$:
\begin{equation}
    \bm{S}= \mathcal {F}_{3}^{1\times1} ({\textstyle \sum_{i=1}^{2}} (\bm{\widetilde{SA}}_{i} \otimes \bm{U}_{i})),
\end{equation}
where $\otimes$ is the element-wise multiplication.

\textbf{Modality Fusion:} 
To obtain the final modality fusion feature $\bm{Z}$, we treat $\bm{S}$ as the attention weights for HR guidance feature $\bm{Y}$ and perform element-wise multiplication, denoted as:
\begin{equation}
    \bm{Z}= \bm{Y} \otimes \bm{S}.
    \label{equ:7}
\end{equation}

\subsubsection{Modality-Conditioned Weight Generation Pathway}\label{subsubsec:MCWG-pathway}

\begin{figure*}[htbp]
  \centering
    \vspace{-1\baselineskip}
  \includegraphics[width=.99\textwidth]{"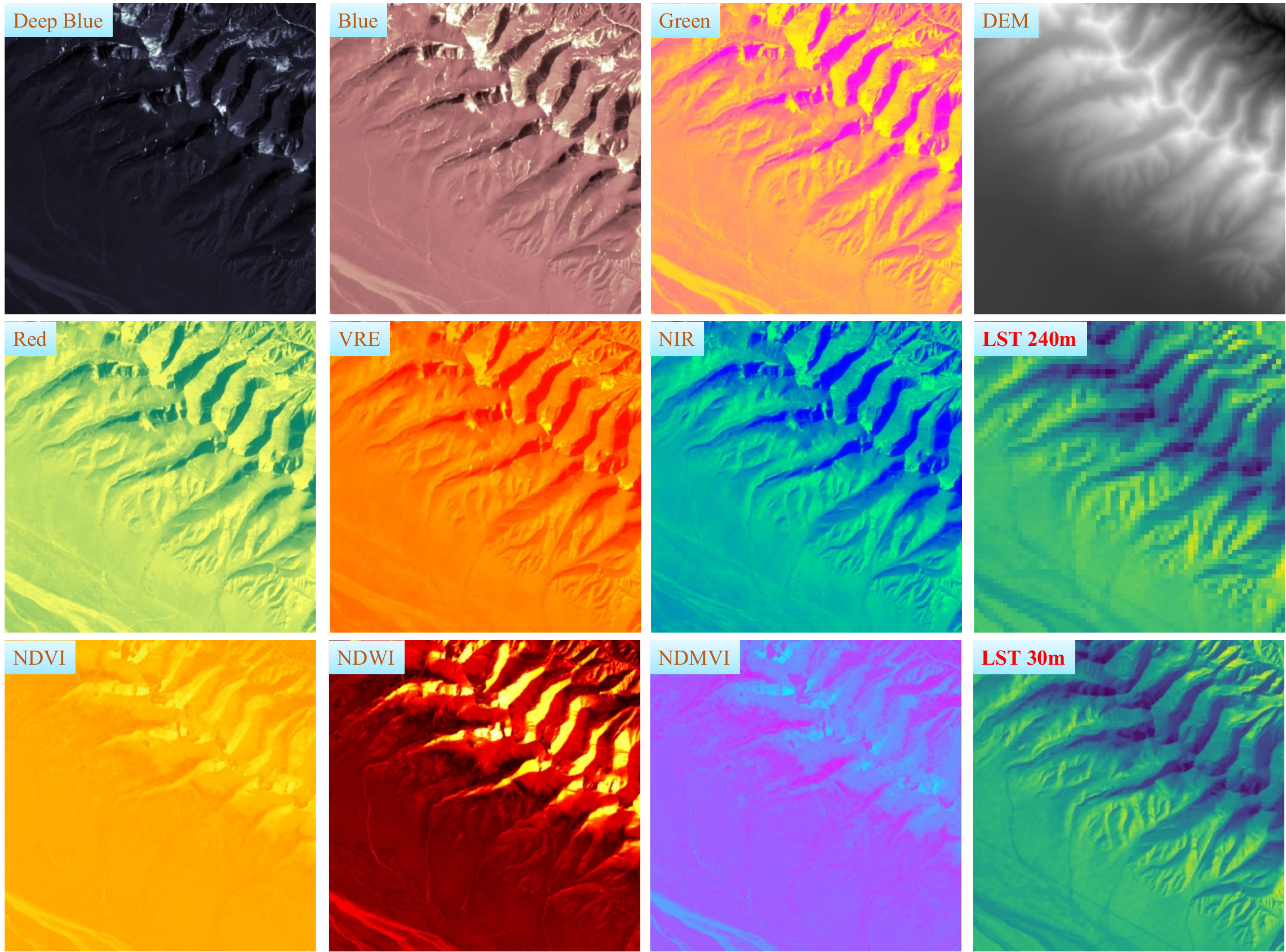"}
      % \vspace{-.5\baselineskip}
  \caption{Gallery of the GrokLST dataset, showcasing the comparison between LR (240 m) and HR (30 m) LST images, along with a suite of 10 auxiliary data.}
  \label{fig:LST-Gallery-3}
  \vspace{-1\baselineskip}
\end{figure*}

The MCWG pathway generates modality-conditioned weights for HR LST feature $\bm{X}$ under the guidance of modality $\bm{Y}$ using a pyramid pooling module (PPM) and a dynamic MLP (DMLP) \cite{CVPR2022DMLP}, which can be formulated as (with reshaping and other operations omitted for clarity):
\begin{equation}
    \begin{aligned}
    	weights  & = \text{MCWG} (\bm{X},\bm{Y}) \\
        	           & = \text{DMLP}(\text{PPM}(\bm{X}), \text{PPM}(\bm{Y})),
    \end{aligned}
\end{equation}
\begin{equation}
    \text{PPM}(\bm{X/Y}) = \left [ AvgPool_{i}(\bm{X/Y}) \right ],\ i=1,2,3,4,
\end{equation}
where $AvgPool_{i}(\cdot)$ represents a series of global average poolings with bin sizes $\left \{ 1,2,3,6 \right \}$, similar to pyramid scene parsing network (PSPNet) \cite{CVPR2017PSP}.

% \subsubsection{Integration and Dynamic Interaction}

% By integrating the LSK pathway and the MCWG pathway, the MoCoLSK module effectively captures multi-scale features and performs adaptive multi-modal fusion. This dynamic adaptation allows the network to leverage the complementary information from the auxiliary data and improve the accuracy of land surface temperature downscaling. The complete MoCoLSK Module can be formulated as:
% \begin{equation}
%     \begin{aligned}
%         T_{lr}^{(l)}  & = \text{MoCoLSK}(T_{lr}^{(l-1)}, G_{hr}^{(l-1)}), \\
%                 & = \text{Down} (\left [ \text{Up}(T_{lr}^{(l-1)}), \bm{Z} \right ]), 
%     \end{aligned}
% \end{equation}
% where $\text{MoCoLSK}$ denotes the complete MoCoLSK module. $T_{lr}^{(l-1)}$ and $G_{hr}^{(l-1)}$, representing the LR LST feature and HR guidance feature, serve as the input features for the MoCoLSK module of stage $l$. Up and Down are up-projection and down-projection units \cite{CVPR2018DBPN}, respectively. You can go to \ref{subsubsec:LSK-pathway} to figure out how $\bm{Z}$ is generated. $T_{lr}^{(l)}$ is the output of the MoCoLSK module at stage $l$.

% !TEX root = ../main.tex
% \bibliography{../reference.bib}

\section{Experiments} \label{sec:experiment}

\subsection{Experimental Settings} \label{subsec:setting}

% \begin{figure*}[htbp]
%   \centering
%     \vspace{-1\baselineskip}
%   \includegraphics[width=.99\textwidth]{"./figs/ecosystem/LST-Gallery-3.pdf"}
%       % \vspace{-.5\baselineskip}
%   \caption{Gallery of the GrokLST dataset, showcasing the comparison between LR (240 m) and HR (30 m) LST images, along with a suite of 10 auxiliary data products.}
%   \label{fig:LST-Gallery-3}
%   % \vspace{-1\baselineskip}
% \end{figure*}

\begin{table}[htbp]
\renewcommand\arraystretch{1}
\vspace{-1\baselineskip}
\caption{The corresponding sizes of LST and guidance at different resolutions in the GrokLST dataset. H: height, W: width, C: channel.} \label{tab:size}
\vspace{2pt}
\centering
\setlength{\tabcolsep}{2pt}

\begin{tabular}{c|c|c|c} \toprule[1pt]
Resolution & Scale & LST Size (H$\times$W$\times$C) & Guidance Size (H$\times$W$\times$C) \\ \midrule[1pt]
30m & - & 512$\times$512$\times$1 & 512$\times$512$\times$10 \\
60m & $\times$2 & 256$\times$256$\times$1 & 256$\times$256$\times$10 \\
120m & $\times$4 & 128$\times$128$\times$1 & 128$\times$128$\times$10 \\
240m & $\times$8 & 64$\times$64$\times$1 & 64$\times$64$\times$10 \\ \bottomrule[1pt]
\end{tabular}
\vspace{-1\baselineskip}
\end{table}

% reviewer 2-comments 2 & 7，添加了细节数据集相关描述以及
\subsubsection{\textbf{Dataset}}\label{subsubsec:V.A.dataset}
% \rewrite{}
% In addressing various downscaling challenges, we have adhered to Wald’s protocol to downsample the data to multiple resolutions, specifically 60, 120, and 240 meters. This approach facilitates a comprehensive exploration of various downscaling factors. 
We utilize our GrokLST dataset for experiments. To address the challenge of downscaling across different resolutions, we adhere to the Wald’s protocol, downsampling the 30m resolution data to three distinct resolutions of 60m, 120m, and 240m, thereby enabling $\times$2, $\times$4, and $\times$8 downscaling tasks. Specifically, the 30m resolution LST data is used as the ground truth (GT), while LST data at other resolutions is downscaled with the aid of 30m guidance data to reconstruct the predicted 30m resolution LST. The specific spatial resolutions of the GrokLST dataset are detailed in Table \ref{tab:size}. Fig.~\ref{fig:LST-Gallery-3} provides visual representations of these bands and indices, highlighting the spectral characteristics and quality of the dataset. 
% For effective model training and validation, the GrokLST dataset is carefully segmented into three subsets: 60\% for training, 10\% for validation, and 30\% for testing. 
For effective model training and validation, the GrokLST dataset is carefully divided into three subsets in a 6:1:3 ratio: 384 samples for training, 64 for validation, and 193 for testing.
All LST and guidance data are processed using the Z-score normalization strategy. For an in-depth analysis of different normalization strategies, please refer to \ref{subsec:norm}.

\subsubsection{\textbf{Evaluation Metrics}}
Some key statistical indicators such as root mean square error (RMSE), mean absolute error (MAE), bias (BIAS), correlation coefficient (CC), and ratio of standard deviations (RSD) are utilized to quantitatively evaluate reconstruction performance of one downscaling model.
% with their specific formulas as follows:

RMSE is the square root of the average of the squared differences between the predicted HR LST $T_{sr}$ and the ground truth $T_{hr}$:
\begin{equation}
RMSE=\sqrt{\frac{1}{N} {\textstyle \sum_{i=1}^{N}}  ( T_{sr}^{i} - T_{hr}^{i} )^{2} }.
\end{equation}

MAE represents the average of the absolute differences between $T_{sr}$ and $T_{hr}$:
\begin{equation}
MAE=\frac{1}{N} {\textstyle \sum_{i=1}^{N}}  \left |  T_{sr}^{i} - T_{hr}^{i} \right |.
\end{equation}

BIAS shows the average of the differences between $T_{sr}$ and $T_{hr}$:
\begin{equation}
BIAS=\frac{1}{N} {\textstyle \sum_{i=1}^{N}}  (T_{sr}^{i} - T_{hr}^{i}).
\end{equation}

CC evaluates the correlation between $T_{sr}$ and $T_{hr}$, with a value of 1 indicating perfect correlation:
\begin{equation}
CC=\frac{\frac{1}{N} {\textstyle \sum_{i=1}^{N}} ( \Delta T_{sr}^{i} )( \Delta T_{hr}^{i} )}{\sqrt{\frac{1}{N} {\textstyle \sum_{i=1}^{N}} ( \Delta T_{sr}^{i} )^{2} } \sqrt{\frac{1}{N} {\textstyle \sum_{i=1}^{N}} ( \Delta T_{hr}^{i} )^{2}}} ,
\end{equation}
where 
\begin{equation}
    \begin{aligned}
    	\Delta T_{sr}^{i} & =T_{sr}^{i} - \frac{1}{N}{\textstyle \sum_{i=1}^{N} T_{sr}^{i}}, \\
            \Delta T_{hr}^{i} & =T_{hr}^{i} - \frac{1}{N}{\textstyle \sum_{i=1}^{N} T_{hr}^{i}}.
    \end{aligned}
\end{equation}

% RSD compares the standard deviation of $T_{sr}$ to the standard deviation of $T_{hr}$:
RSD quantifies how closely the distribution of $T_{sr}$ matches the distribution of $T_{hr}$:
\begin{equation}
RSD=\frac{ \left | \sigma_{sr}- \sigma_{hr} \right | }{\sigma_{hr}},
\end{equation}
where
\begin{equation}
    \begin{aligned}
    	\sigma_{sr} & =\sqrt{\frac{1}{N-1} {\textstyle \sum_{i=1}^{N}} ( T_{sr}^{i} - \overline{T}_{sr} )^{2}}, \\
            \sigma_{hr} & =\sqrt{\frac{1}{N-1} {\textstyle \sum_{i=1}^{N}} ( T_{hr}^{i} - \overline{T}_{hr} )^{2}}.
    \end{aligned}
\end{equation}
% $N$ is the number of LST images in the HeiheLST dataset. 

The closer the values of RMSE, MAE, BIAS, and RSD are to 0, and the closer CC is to 1, the better the downscaling method's reconstruction performance.

\subsubsection{\textbf{Implementation Details}}\label{sebsebsec:5.1.2}

We implemented our MoCoLSK-Net in our GrokLST toolkits and trained it on a platform equipped with four NVIDIA GeForce RTX 4090 GPUs using a distributed training approach. 
During training, we employ the AdamW optimizer \cite{arXiv2017AdamW} and a cosine annealing learning rate scheduler with warm restarts \cite{arXiv2016sgdr}. 
The initial learning rate is set to $1e\text{-}4$ and weight decay by a factor of $1e\text{-}5$ for $10 k$ iterations. Each GPU is assigned one training sample, and the batch size is fixed to 4. All other deep learning-based methods in the GrokLST toolkits use the above experimental configuration. 

Next, we detail the key hyperparameters in MoCoLSK-Net. 

1) \textbf{Base Feature Dimension:} In the guidance branch, the feature dimension of all residual groups remains fixed at 32. In contrast, in the LST branch, the feature dimension of the residual groups increases by 32 in each stage compared to the previous stage. Besides, the feature dimension of all submodules within the reconstruction module is maintained at $N\times32$. 

2) \textbf{Number of Stages:} MoCoLSK-Net by default has 4 stages (i.e., $N=4$), with each stage containing two residual groups and one MoCoLSK module. 

3) \textbf{Number of Layers in DMLP:} The DMLP contains multiple linear layers to enhance the dynamic fitting capability of the module. In MoCoLSK-Net, the default number of DMLP layers is 1. 

4) \textbf{DMLP Versions:} There are three versions of standard DMLP, namely A, B, and C. For details, please refer to \cite{CVPR2022DMLP}.

5) \textbf{Size of Weights:} The weights dynamically generated by MCWG pathway are used in DConv in LSK pathway to obtain modality-conditioned spatial selection masks. The default size of the weights is 3$\times$3.

% 对比实验
\begin{table*}[htbp]
  \renewcommand\arraystretch{1.2}
  \footnotesize
  \centering
  \vspace{-1\baselineskip}
  \caption{Comparison with state-of-the-art methods on the \textbf{GrokLST} dataset. The symbol “-” indicates insufficient memory to execute the algorithm, while “\ding{55}” denotes that the algorithm does not support the corresponding downscaling factor.}\label{tab:sota}
  \vspace{-2pt}
  \setlength{\tabcolsep}{1.5pt}
  
  \begin{tabular}{l|ccccc|ccccc|ccccc}
  \multirow{2}{*}{Method} & \multicolumn{5}{c|}{$\times 2$} & \multicolumn{5}{c|}{$\times 4$} & \multicolumn{5}{c}{$\times 8$} \\
  & RMSE$\downarrow$ & MAE$\downarrow$ & BIAS & CC$\uparrow$ & RSD$\downarrow$ 
  & RMSE$\downarrow$ & MAE$\downarrow$ & BIAS & CC$\uparrow$ & RSD$\downarrow$ 
  & RMSE$\downarrow$ & MAE$\downarrow$ & BIAS & CC$\uparrow$ & RSD$\downarrow$  \\
  \Xhline{1pt}
  \multicolumn{16}{l}{\textit{Machine Learning}}  \\ \hline
    % Bilinear & 0.6338 & 0.4292 & 0.0000  & 0.9794 & 0.0640 & 1.1798 & 0.8175 & 0.0000  & 0.9375 & 0.1277 & 1.7893 & 1.2787 & 0.0000 & 0.8664 & 0.2130 \\
    % Bicubic  & 0.4818 & 0.3204 & 0.0000  & 0.9859 & 0.0278 & 1.0323 & 0.7069 & 0.0000  & 0.9478 & 0.0819 & 1.6709 & 1.1811 & 0.0000 & 0.8788 & 0.1612 \\
    Random Forest \cite{ML2001RF} & - & - & - & - & - & 1.3900 & 0.9494 & -0.0317 & 0.9055 & 0.0712 & 1.7367 & 1.2264 & -0.0456 & 0.8477 & 0.1330 \\
    XGBoost \cite{SIGKDD2016xgboost} & - & - & - & - & - & 1.8209 & 1.3000 & -0.0108 & 0.8244 & 0.1228 & 1.9825 & 1.4342 & -0.0156 & 0.7846 & 0.1755 \\
    LightGBM \cite{NIPS2017lightgbm} & - & - & - & - & - & 1.7826 & 1.2680 & -0.0202 & 0.8340 & 0.1016 & 2.0205 & 1.4632 & -0.0267 & 0.7772 & 0.1425 \\
    CatBoost \cite{NIPS2018catboost} & - & - & - & - & - & 1.5350 & 1.0639 & -0.0175 & 0.8787 & 0.0866 & 1.9089 & 1.3683 & -0.0074 & 0.8030 & 0.1537 \\
  \hline
  \multicolumn{16}{l}{\textit{Single Image Downscaling}}  \\
  \hline
    % SRCNN    & 0.6793 & 0.3652 & -0.0048 & 0.9663 & 0.0182 & 1.0735 & 0.6949 & 0.0007  & 0.9366 & 0.0455 & 1.6567 & 1.1582 & 0.0297  & 0.8722 & 0.1100 \\
    EDSR \cite{lim2017enhanced}     & 0.4010 & 0.2605 & 0.0018  & 0.9889 & 0.0114 & 0.8921 & 0.6042 & 0.0061  & 0.9559 & 0.0441 & 1.4855 & 1.0397 & 0.0112  & 0.8933 & 0.1024 \\
    RDN \cite{CVPR2018RDN}      & 0.3802 & 0.2478 & 0.0018  & 0.9898 & 0.0104 & 0.8227 & 0.5598 & 0.0066  & 0.9595 & 0.0400 & 1.2497 & 0.8742 & 0.0133  & 0.9110 & 0.0856 \\
    RCAN \cite{zhang2018image}     & 0.4046 & 0.2644 & 0.0017  & 0.9887 & 0.0117 & 0.8826 & 0.6009 & 0.0065  & 0.9562 & 0.0440 & 1.4446 & 1.0147 & 0.0114  & 0.8958 & 0.1017 \\
    DBPN \cite{CVPR2018DBPN}     & 0.4257 & 0.2803 & 0.0008  & 0.9879 & 0.0120 & 0.8865 & 0.6008 & 0.0072  & 0.9564 & 0.0431 & 1.4303 & 0.9982 & 0.0146  & 0.8975 & 0.0991 \\
    CTNet \cite{TGRS2021CTNet}      & 0.4012 & 0.2627 & 0.0021  & 0.9889 & 0.0118 & 0.8954 & 0.6064 & 0.0065  & 0.9561 & 0.0442 & \ding{55}      & \ding{55}      & \ding{55}       & \ding{55}      & \ding{55}      \\
    FeNet \cite{TGRS2022FeNet}    & 0.3995 & 0.2616 & 0.0017  & 0.9890 & 0.0116 & 0.9009 & 0.6118 & 0.0064  & 0.9557 & 0.0457 & 1.5193 & 1.0644 & 0.0113  & 0.8917 & 0.1074 \\
    FENet \cite{IEEEAccess2022FENet}    & 0.4018 & 0.2632 & 0.0020  & 0.9889 & 0.0117 & 0.8879 & 0.6017 & 0.0070  & 0.9564 & 0.0440 & 1.4844 & 1.0366 & 0.0135  & 0.8940 & 0.1030 \\
    SRFBN \cite{CVPR2019srfbn}    & 0.3962 & 0.2586 & 0.0020  & 0.9891 & 0.0113 & 0.8969 & 0.6067 & 0.0073  & 0.9560 & 0.0446 & 1.5212 & 1.0589 & 0.0131  & 0.8919 & 0.1060 \\
    CFGN \cite{TETCI2024CFGN}    & 0.4283 & 0.2814 & 0.0014  & 0.9877 & 0.0127 & 0.9386 & 0.6395 & 0.0053  & 0.9529 & 0.0485 & 1.5724 & 1.1052 & 0.0108  & 0.8867 & 0.1134 \\
    \textit{SwinIR} \cite{ICCVW2021SwinIR} & - & - & - & - & - & 0.9259 & 0.6297 & 0.0059 & 0.9537 & 0.0474 & 1.5549 & 1.0906 & 0.0104 & 0.8884 & 0.1112 \\
    \textit{DAT} \cite{ICCV2023DAT} & - & - & - & - & - & - & - & - & - & - & 1.2605 & 0.8816 & 0.0153 & 0.9131 & 0.0842 \\
    \textit{SRFormer} \cite{ICCV2023SRFormer} & - & - & - & - & - & 0.9203 & 0.6252 & 0.0065 & 0.9544 & 0.0473 & 1.5584 & 1.0929 & 0.0123 & 0.8882 & 0.1122 \\
    \textit{DLGSANet} \cite{ICCV2023DLGSANet} & - & - & - & - & - & 0.9559 & 0.6537 & 0.0059 & 0.9514 & 0.0483 & 1.5848 & 1.1159 & 0.0097 & 0.8852 & 0.1128 \\
    \textit{ACT} \cite{WACV2023ACT} & - & - & - & - & - & 0.8908 & 0.6054 & 0.0057 & 0.9560 & 0.0444 & 1.4592 & 1.0265 & 0.0117 & 0.8952 & 0.1021 \\
    \textit{NGSwin} \cite{CVPR2023NGram} & 0.4902 & 0.2897 & 0.0006 & 0.9831 & 0.0113 & 1.3168 & 0.6786 & 0.0017 & 0.8904 & 0.0694 & \ding{55} & \ding{55} & \ding{55} & \ding{55} & \ding{55} \\
    \textit{DCTLSA}  \cite{TIM2023DCTLSA} & 0.4616 & 0.2981 & -0.0056 & 0.9855 & 0.0226 & 0.9837 & 0.6720 & 0.0212 & 0.9456 & 0.0590 & 1.6282 & 1.1488 & -0.0107 & 0.8708 & 0.1156 \\
    \textit{HiT-SIR} \cite{ECCV2025HiT-SR} & 0.4078 & 0.2675 & 0.0023 & 0.9886 & 0.0117 & 0.9009 & 0.6121 & 0.0067 & 0.9555 & 0.0444 & 1.5301 & 1.0735 & 0.0132 & 0.8904 & 0.1066 \\
    \textit{HiT-SNG} \cite{ECCV2025HiT-SR} & 0.4084 & 0.2677 & 0.0014 & 0.9886 & 0.0118  & 0.9079 & 0.6176 & 0.0056 & 0.9551 & 0.0453 & 1.5207 & 1.0669 & 0.0123 & 0.8912 & 0.1057 \\
    \textit{HiT-SRF} \cite{ECCV2025HiT-SR} & 0.4080 & 0.2674 & 0.0018 & 0.9886 & 0.0116 & 0.9028 & 0.6136 & 0.0064 & 0.9554 & 0.0448 & 1.5142 & 1.0622 & 0.0092 & 0.8916 & 0.1049 \\
  \hline  				
  \multicolumn{16}{l}{\textit{Guided Image Downscaling}}  \\
  \hline
    MSG-Net \cite{ECCV2016MSGNet}      & 0.4294 & 0.2829 & 0.0021  & 0.9877 & 0.0120 & 0.8651 & 0.5914 & 0.0043  & 0.9578 & 0.0412 & 1.3418 & 0.9442 & 0.0070  & 0.9048 & 0.0893 \\
    % DJF      & 1.1447 & 0.5233 & -0.0023 & 0.9074 & 0.0738 & 1.1813 & 0.6949 & 0.0273  & 0.9059 & 0.0629 & 1.3644 & 0.9512 & 0.0156  & 0.8897 & 0.0759 \\
    % DGF \cite{CVPR2018DGF}      & 0.6577 & 0.4447 & 0.0000  & 0.9767 & 0.0571 & 1.0702 & 0.7339 & 0.0003  & 0.9448 & 0.0907 & 1.6574 & 1.1707 & 0.0002  & 0.8795 & 0.1589 \\
    SVLRM \cite{CVPR2019SVLRM}    & 0.4612 & 0.3085 & 0.0045  & 0.9863 & 0.0232 & 0.8611 & 0.5974 & 0.0017  & 0.9567 & 0.0513 & 1.2357 & 0.8815 & -0.0101 & 0.9135 & 0.0863 \\
    DJFR \cite{TPAMI2019DJFR}     & 0.3933 & 0.2603 & 0.0013  & 0.9891 & 0.0112 & 0.7784 & 0.5382 & 0.0031  & 0.9642 & 0.0390 & 1.1892 & 0.8436 & 0.0010  & 0.9181 & 0.0809 \\
    P2P \cite{CVPR2019P2P}      & 0.4788 & 0.3200 & -0.0041 & 0.9860 & 0.0227 & 1.0003 & 0.6898 & -0.0298 & 0.9497 & 0.0602 & 1.5409 & 1.0952 & -0.0964 & 0.8914 & 0.1062 \\
    DSRN \cite{CVPR2018DSRN}     & 0.4480 & 0.2956 & 0.0574  & 0.9875 & 0.0243 & 0.9562 & 0.6543 & 0.1458  & 0.9525 & 0.0724 & 1.5587 & 1.1023 & 0.2025  & 0.8891 & 0.1477 \\
    FDSR \cite{he2021towards}     & 0.4065 & 0.2698 & 0.0013  & 0.9885 & 0.0125 & 0.7779 & 0.5371 & 0.0047  & 0.9619 & 0.0396 & 1.1395 & 0.8096 & 0.0047  & 0.9210 & 0.0783 \\
    DKN \cite{IJCV2021DKNFDKN}      & 0.4071 & 0.2695 & 0.0041  & 0.9884 & 0.0135 & 0.8388 & 0.5727 & 0.0026  & 0.9574 & 0.0416 & 1.3719 & 0.9589 & 0.0036  & 0.8976 & 0.1042 \\
    FDKN \cite{IJCV2021DKNFDKN}     & 0.3717 & 0.2449 & 0.0032  & 0.9901 & 0.0099 & 0.7946 & 0.5456 & 0.0032  & 0.9612 & 0.0387 & 1.3312 & 0.9335 & 0.0038  & 0.9061 & 0.0978 \\
    AHMF \cite{zhong2021high}     & 0.3557 & 0.2348 & 0.0017  & 0.9908 & 0.0097 & 0.7224 & 0.4996 & 0.0028  & 0.9655 & 0.0352 & 1.1246 & 0.7959 & 0.0019  & 0.9229 & 0.0764 \\
    CODON \cite{IJCV2022CODON}    & \ding{55}      & \ding{55}      & \ding{55}       & \ding{55}      & \ding{55}      & 0.9617 & 0.6642 & 0.0210  & 0.9492 & 0.0531 & 1.6690 & 1.1799 & -0.0001 & 0.8789 & 0.1608 \\
    SUFT \cite{shi2022symmetric}     & \ul{0.3130} & \ul{0.2093} & 0.0011  & \ul{0.9927} & \ul{0.0075} & \ul{0.6046} & \ul{0.4207} & 0.0021  & \ul{0.9737} & \ul{0.0265} & \ul{0.8598} & \ul{0.6061} & 0.0025  & \ul{0.9468} & \ul{0.0489} \\
    DAGF \cite{TNNLS2023DAGF}     & 0.3917 & 0.2589 & 0.0005  & 0.9892 & 0.0110 & 0.7910 & 0.5451 & 0.0063  & 0.9613 & 0.0391 & 1.1935 & 0.8469 & 0.0070  & 0.9170 & 0.0879 \\
    RSAG \cite{yuan2023recurrent}     & \ding{55}      & \ding{55}      & \ding{55}       & \ding{55}      & \ding{55}      & 0.7223 & 0.4990 & 0.0026  & 0.9654 & 0.0350 & 1.0118 & 0.7154 & 0.0012  & 0.9330 & 0.0638 \\
  \hline
  	 	 	 	 
  \rowcolor[rgb]{0.9,0.9,0.9}$\star$ \textbf{MoCoLSK}  
  & \textbf{0.2902} & \textbf{0.1951} & 0.0009  & \textbf{0.9937} & \textbf{0.0062}
  & \textbf{0.5590} & \textbf{0.3883} & 0.0020  & \textbf{0.9771} & \textbf{0.0218} 
  & \textbf{0.8031} & \textbf{0.5642} & 0.0027  & \textbf{0.9514} & \textbf{0.0456}  \\
  \end{tabular}
  \vspace{-1\baselineskip}
\end{table*}

\subsection{Comparison with State-of-the-Arts} \label{subsec:sota}

We benchmarked the reconstruction performance of MoCoLSK-Net against current state-of-the-art downscaling methods, including four machine learning methods, nineteen single-image downscaling methods, and thirteen guided image downscaling methods. The benchmarking covers three scales: $\times$2, $\times$4, and $\times$8, with evaluation conducted on our GrokLST dataset using five metrics: RMSE, MAE, BIAS, CC, and RSD. All comparison methods and experimental results are listed in Table \ref{tab:sota}, and the following key conclusions can be drawn:

1) Our MoCoLSK-Net achieves state-of-the-art performance across nearly all metrics (except BIAS) in downscaling challenges at various scales, underscoring the effectiveness of MoCoLSK, which utilizes modality-conditioned dynamic receptive fields for multimodal fusion. The MoCoLSK module integrates the LSK and MCWG pathways, working synergistically to enable deep dynamic fusion of LST and guidance features, continuously refining the discriminative LST features. As a result, MoCoLSK-Net outperforms all other SOTA methods, delivering the most accurate LST downscaling results.

2) All deep learning-based downscaling methods significantly outperformed the four traditional machine learning methods (Random Forest \cite{ML2001RF}, XGBoost \cite{SIGKDD2016xgboost}, LightGBM \cite{NIPS2017lightgbm}, and CatBoost \cite{NIPS2018catboost}) across all reconstruction scales. 
This indicates that the four traditional machine learning methods struggle to accurately capture the fine-grained mapping relationships between multiple guidance data and LST. This may be due to the fact that these models learn the global mapping between LST and the multiple guidance data, and then predict LST on a pixel-by-pixel basis \cite{RSE2022nonlineardownscaling}. Although LST is highly correlated with surface attributes in the guidance data, this relationship may vary across different regions, and the global mapping may not fully satisfy the local downscaling of LST. Furthermore, the pixel-wise reconstruction process can disrupt the spatial texture of LST, resulting in noticeable temperature biases, either very high or very low, in the downscaled LST.

3) Among single-image downscaling methods, we did not observe Transformer-based downscaling algorithms (italicized in Table \ref{tab:sota}) outperforming those based on CNNs, spatial attention, or channel attention across various metrics. Instead, Transformer-based methods introduced greater computational complexity, as evidenced by the significant number of “-” entries in Table \ref{tab:sota}. This suggests that algorithms leveraging CNNs, spatial, or channel attention mechanisms are not inferior to Transformer methods with global attention. It is worth noting, however, that the relatively small size of our GrokLST dataset might limit the reconstruction performance of Transformer-based methods.

4) Overall, guided downscaling methods exhibited significantly better performance across reconstruction metrics compared to single-image downscaling methods. Examples include FDKN, AHMF, SUFT, and RSAG, with our MoCoLSK significantly outperforming RDN (the most effective single-image downscaling algorithm). This highlights not only the higher reconstruction quality ceiling of guided downscaling methods compared to single-image methods but also underscores the importance and necessity of HR guided data.

\begin{figure*}[h]
    \centering
    \includegraphics[width=1\textwidth]{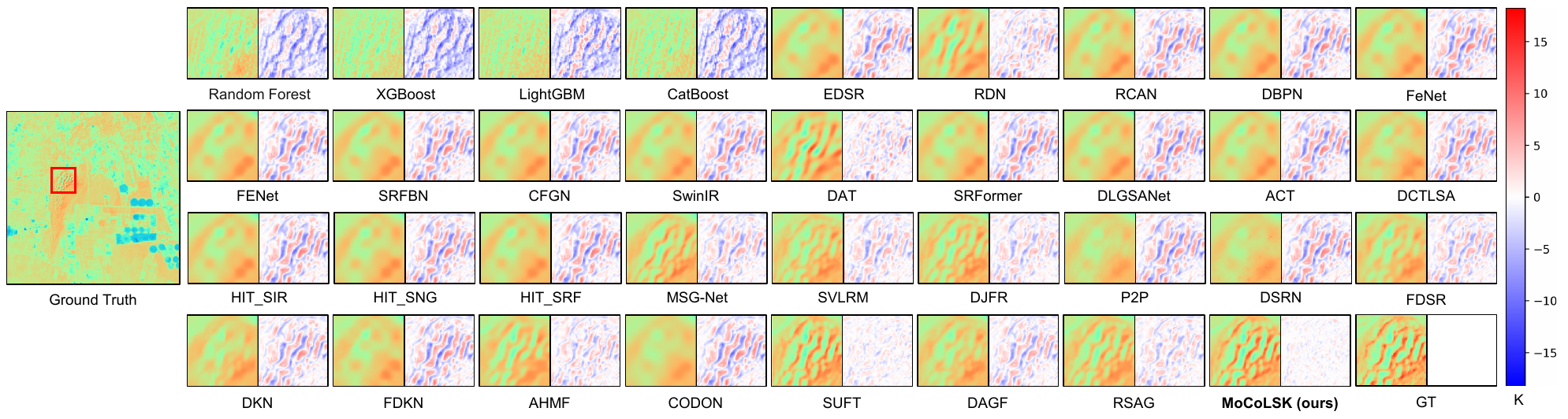}
    \caption{Visual comparison of $\times$8 reconstruction images from different methods. Each method is represented by two images: the left image shows the reconstruction result, while the right image illustrates the difference between the reconstruction result and the GT. A Kelvin (K) temperature color bar is shown on the far right, where pixels with values larger than the GT are displayed in red, smaller values are displayed in blue, and identical values are shown in white.
    }\label{fig:visualization}
\end{figure*}

\subsection{Visual Analysis} \label{subsec:visualization}

Fig. \ref{fig:visualization} provides intuitive visualizations of different downscaling methods on GrokLST dataset at $\times$8 downscaling challenge, with each method showing its downscaled result and difference map compared to GT, complementing quantitative analysis from a qualitative perspective. From these visualizations, we can intuitively draw the following crucial insights:

1) From downscaled result maps, MoCoLSK-Net demonstrates the clearest land surface textures and most accurate temperature predictions. From difference maps, it is evident that MoCoLSK-Net’s difference map aligns most closely with label differences, tending towards white (whiter difference maps indicate better reconstruction performance). This once again offers a comprehensive and intuitive qualitative validation of MoCoLSK-Net's superior LST downscaling performance.

2) Downscaling results of four traditional machine learning algorithms perform the worst visually. These methods fail to reconstruct land surface textures and structures, appearing disordered and lacking smoothness, while showing significant temperature bias. This suggests difficulty in accurately capturing fine-grained mapping between guide data and LST, visually reinforcing significant limitations of paradigms that learn global mappings between LST and multiple guide data for point-by-point LST prediction.

3) In single-image downscaling methods, most algorithms produce blurred LST results with unclear textures and noticeable temperature biases, whereas RDN and DAT yield relatively more accurate LST downscaling. Furthermore, most Transformer-based algorithms (excluding DAT), despite global receptive fields, do not outperform CNN-based or spatial/channel attention-based methods and introduce higher computational complexity. This suggests that for downscaling tasks, CNN-based or attention mechanism-based methods perform similarly to Transformer methods with global receptive fields, further confirmed from a visual perspective.

4) Most guided downscaling methods (e.g., AHMF, SUFT, RSAG) exhibit significantly better reconstruction results than single-image downscaling methods, especially MoCoLSK. This not only highlights superior reconstruction potential of guide-based methods but also further validates the critical role of HR guide data in improving LST downscaling performance.

\section{Disscussion} \label{sec:experiment}

\subsection{Ablation Study} \label{subsec:ablation}
This section presents ablation study results for key components of MoCoLSK module on GrokLST dataset with $\times$8 downscaling and 20k iterations to ensure more reliable results, including large kernel decomposition (LKD), DConv, modality fusion (MF) in LSK pathway, and PPM in MCWG pathway.

\begin{table}[]
\renewcommand\arraystretch{1.2}
\vspace{-1\baselineskip}
\caption{Validation of key components in MoCoLSK module. } \label{tab:2-component}
\vspace{2pt}
\centering
\setlength{\tabcolsep}{2.2pt}
\begin{tabular}{c|ccc|cc|cc}
\multicolumn{1}{c|}{\multirow{2}{*}{Case}} & \multicolumn{3}{c|}{LSK Pathway}            & \multicolumn{2}{c|}{MCWG pathway}   & \multicolumn{2}{c}{Metrics} \\
\multicolumn{1}{c|}{}                     & LKD & DConv & \multicolumn{1}{c|}{MF} & PPM & \multicolumn{1}{c|}{AvgMax} & RMSE$\downarrow$          & CC$\uparrow$  \\ \midrule[1pt]
1 & -   & -     & -  & -   & -      & 0.7405          & 0.9605          \\
2 &     & \checkmark     & \checkmark  & \checkmark   &        & 0.7154          & 0.9612          \\
3 & \checkmark   &       & \checkmark  & \checkmark   &        & 0.7267          & 0.9603          \\
4 & \checkmark   & \checkmark     &    & \checkmark   &        & 0.9407          & 0.9414          \\
5 & \checkmark   & \checkmark     & \checkmark  &     & \checkmark      & 0.7153          & 0.9613          \\
6 & \checkmark   & \checkmark     & \checkmark  & \checkmark   &        & \textbf{0.7133} & \textbf{0.9613} \\ 
\end{tabular}
% \vspace{-1\baselineskip}
\end{table}

Table \ref{tab:2-component} presents the results of the ablation experiments. 
Case 1 is the baseline, which only uses up and down projection layers in MoCoLSK without utilizing the LSK and MCWG pathways, as shown in Fig. \ref{fig:multimodal-sm}(a).
``LKD'' indicates whether large kernel decomposition is utilized; if unchecked, it signifies the use of one large kernel depth-wise convolution with same receptive field (i.e., 23) to replace two decomposed large kernels. ``DConv'' refers to dynamic modality-conditioned convolution; if unchecked, original static depth-wise convolution in LSK is employed. ``MF'' denotes modal fusion, as represented in Equation (\ref{equ:7}); if unchecked, it implements $\bm{Z}= \bm{X} \otimes \bm{S}$, similar to the original LSK.

Cases 2 and 6 validate that large kernel decomposition is superior to a single larger kernel. 
Cases 3 and 6 show that DConv driven by the modality-conditional weights generated by the MCWG pathway performs better than the original static convolution.
Cases 4 and 6 demonstrate the necessity of further modality fusion.
Cases 5 and 6 validate the effectiveness of proposed PPM.

\begin{table}[]
\renewcommand\arraystretch{1.2}
% \vspace{-1\baselineskip}
\caption{Ablation study on the effectiveness of PPM in  MCWG.} \label{tab:4-ppm}
\vspace{2pt}
\centering

\begin{tabular}{ccc|cc}
\multicolumn{3}{c|}{Pooling}                                                       & \multicolumn{2}{c}{Metrics}       \\
Avg.                       & Max.                       & \multicolumn{1}{c|}{PPM}                       & RMSE$\downarrow$            & CC$\uparrow$              \\ \midrule[1pt]
\checkmark &                           &                           & 0.7175          & 0.9610          \\
                          & \checkmark &                           & 0.7237          & 0.9608          \\
\checkmark & \checkmark &                           & 0.7153          & 0.9613          \\
                          &                           & \checkmark & \textbf{0.7133} & \textbf{0.9613}
\end{tabular}
\vspace{-1\baselineskip}
\end{table}

\subsection{Hyperparameters Analysis}

\subsubsection{Pooling in MCWG Pathway}
We conduct an in-depth exploration of different poolings in MCWG pathway. Table \ref{tab:4-ppm} presents the comparative results of different poolings. 
It can be observed that: 
\textbf{1)} Average pooling performs better than max pooling; 
\textbf{2)} Using both average and max pooling together is more effective than using average or max pooling alone; 
\textbf{3)} Our proposed PPM outperforms the other three pooling strategies.

\subsubsection{Base Feature Dimension}

In MoCoLSK-Net, the base feature dimension is also a critical hyperparameter. As the overall channel dimension of MoCoLSK-Net increases, its reconstruction performance naturally improves, as shown in Table \ref{tab:6-dim}. Due to memory limitations, we increased the base dimension only up to 40, but we believe that further appropriate increases in dimension would lead to even better reconstruction performance.

\begin{table}[]
\renewcommand\arraystretch{1.2}
\vspace{-1\baselineskip}
\caption{Study on the impact of base feature dimension on reconstruction performance in MoCoLSK-Net.} \label{tab:6-dim}
\vspace{2pt}
\centering
\setlength{\tabcolsep}{5pt}

\begin{tabular}{c|ccccc}
Dim. & RMSE$\downarrow$            & MAE$\downarrow$             & BIAS   & CC$\uparrow$              & RSD$\downarrow$             \\ \midrule[1pt]
16        & 0.8719          & 0.6175          & 0.0030 & 0.9446          & 0.0512          \\
24        & 0.7932          & 0.5549          & 0.0030 & 0.9536          & 0.0407          \\
32        & 0.7133          & 0.4849          & 0.0038 & 0.9613          & 0.0312          \\
40        & \textbf{0.6697} & \textbf{0.4445} & 0.0037 & \textbf{0.9663} & \textbf{0.0259}
\end{tabular}
\end{table}

\begin{table}[]
\renewcommand\arraystretch{1.2}
\vspace{-1\baselineskip}
\caption{Study on the impact of stage count on reconstruction performance in MoCoLSK-Net.} \label{tab:5-stage}
\vspace{2pt}
\centering
\setlength{\tabcolsep}{5pt}

\begin{tabular}{c|ccccc}
Stages & RMSE$\downarrow$            & MAE$\downarrow$             & BIAS   & CC$\uparrow$              & RSD$\downarrow$             \\ \midrule[1pt]
1      & 0.9076          & 0.6449          & 0.0017 & 0.9413          & 0.0553          \\
2      & 0.8146          & 0.5738          & 0.0030 & 0.9506          & 0.0445          \\
3      & 0.7511          & 0.5200          & 0.0041 & 0.9575          & 0.0360          \\
4      & \textbf{0.7133} & \textbf{0.4849} & 0.0038 & \textbf{0.9613} & 0.0312          \\
5      & 0.7441          & 0.4957          & 0.0042 & 0.9605          & \textbf{0.0306}
\end{tabular}
\vspace{-1\baselineskip}
\end{table}

\subsubsection{Number of Stages}
Table \ref{tab:5-stage} presents the impact of stacking different numbers of residual groups and the MoCoLSK module on LST reconstruction performance. Notably, a stage number of 4 achieves optimal reconstruction performance, whereas a number of 5 leads to a decline in performance.

\subsubsection{Number of Layers in Different Versions of DMLP}
\begin{figure}[t]
    \centering
    \includegraphics[width=.98\columnwidth]{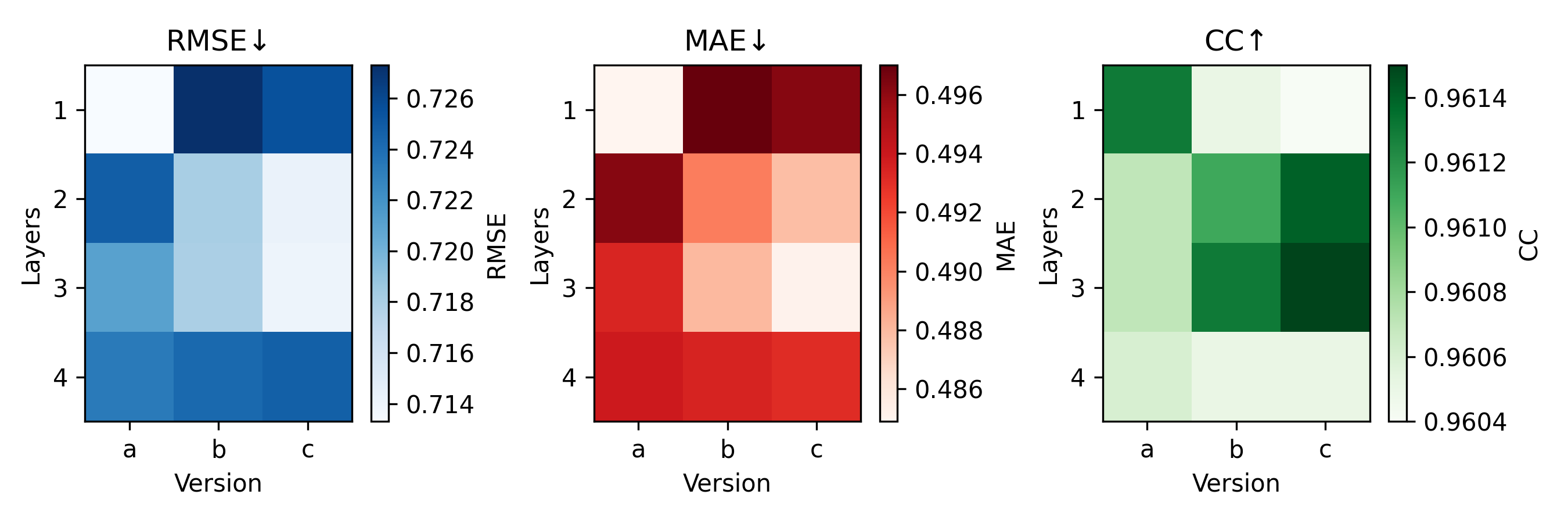}
    \caption{Heatmap of the impact of different layer numbers for the three versions of dynamic MLP (i.e., A, B, and C) on reconstruction performance in the MCWG pathway.}
    \label{fig:layer-in-DMLP}
\end{figure}

The depth of the linear layers in different versions of DMLP determines the quality of modality-conditioned weights, which subsequently affects the LST downscaling performance of MoCoLSK-Net. Fig. \ref{fig:layer-in-DMLP} illustrates the reconstruction performance, as reflected by RMSE, MAE, and CC metrics, for three DMLP versions (A, B, and C) with varying numbers of linear layers. It can be observed that version A of DMLP achieves optimal performance with 1 layer, whereas versions B and C achieve the best reconstruction with 3 layers.

\begin{figure}[ht]
    \centering
    \includegraphics[width=.98\columnwidth]{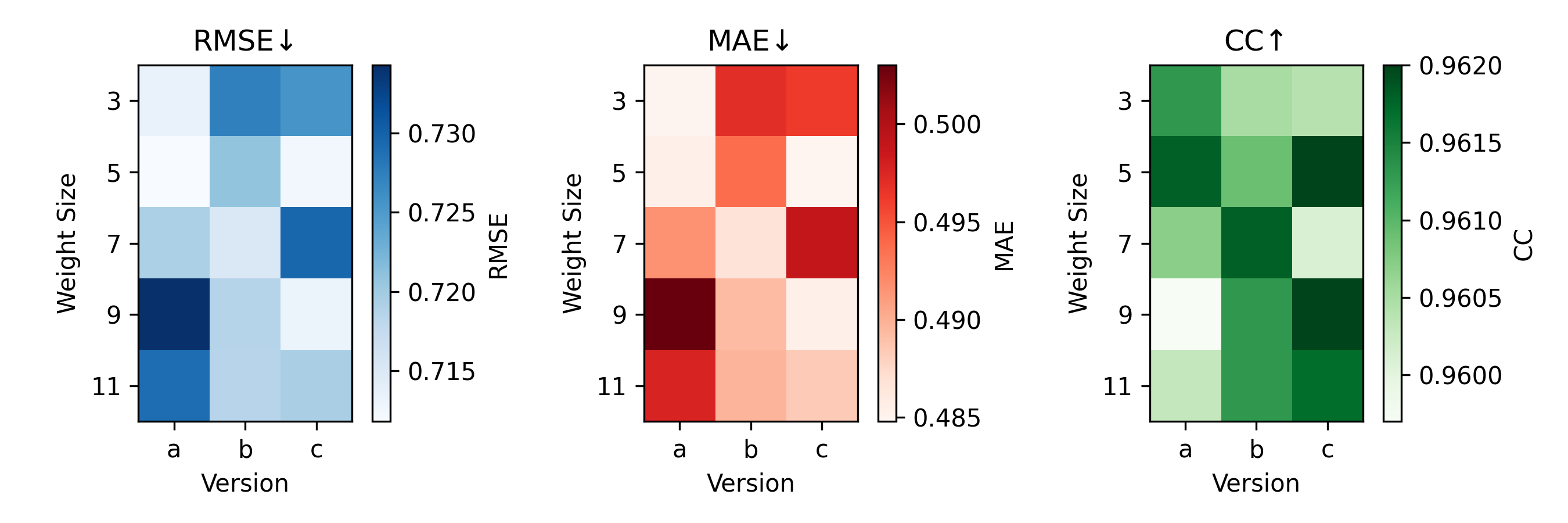}
    \caption{Heatmap of experimental results on reconstruction performance with different weight sizes from three versions of dynamic MLP, \textit{i.e.}, A, B, and C.}
    \label{fig:weight-size}
    % \vspace{-1\baselineskip}
\end{figure}

\subsubsection{Size of Weights}
We conducted a study of how different sizes of modality-conditioned weights, generated by three versions of DMLP, affect LST reconstruction performance. 
Fig. \ref{fig:weight-size} provides detailed experimental results, revealing the following:
the optimal weight size varies across DMLP versions but remains broadly consistent. For version A, the best size is 5$\times$5 (instead of the default 3$\times$3); for version B, it is 7$\times$7; and for version C, 5$\times$5, slightly outperforming 9$\times$9. Moreover, the optimal weight sizes, typically around 5$\times$5 or 7$\times$7, indicate that larger weights do not necessarily lead to more accurate LST downscaling predictions.

\begin{table}[htbp]
\renewcommand\arraystretch{1}
\vspace{-1\baselineskip}
\caption{Research on large kernel convolutions with different receptive fields. Note that the \#P and FLOPs columns only focus on the single MoCoLSK module (ignoring the up and down projection units). K: kernel, D: dilation, RF: receptive field.} \label{tab:3-kernel}
\vspace{2pt}
\centering
\setlength{\tabcolsep}{2pt}

\begin{tabular}{l|ccc|cc}
(K, D) Sequences          & RF & \#P & FLOPs & RMSE$\downarrow$   & CC$\uparrow$     \\  \midrule[1pt]
(23, 1)                      & 23 &  0.02M   & 4.44G      & 0.7154 & 0.9612 \\
(3, 1) $\rightarrow$ (3, 2)  & 7  &  0.04M   & 0.56G      & 0.7233 & 0.9609 \\
(3, 1) $\rightarrow$ (5, 2)  & 11 &  0.04M   & 0.69G      & 0.7183 & 0.9613 \\
(5, 1) $\rightarrow$ (7, 3)  & 23 &  0.04M   & 1.03G      & 0.7133 & 0.9613 \\
(7, 1) $\rightarrow$ (9, 4)  & 39 &  0.04M   & 1.49G      & \textbf{0.7091} & 0.9620 \\
(9, 1) $\rightarrow$ (11, 5) & 59 &  0.05M   & 2.10G      & 0.7092 & \textbf{0.9622} \\
\end{tabular}
\end{table}

\subsection{Larger Kernel, Better Performance?}
We conducted a deeper investigation into large-kernel decomposition to determine whether larger kernels result in better reconstruction performance. Table \ref{tab:3-kernel} shows the impact of a single large kernel and a series of two consecutive large kernels with varying receptive fields on LST downscaling performance. The results reveal the following: a single large kernel convolution with same receptive field 23 is less effective than two consecutive decomposed large-kernel convolutions. Furthermore, the LST downscaling performance of MoCoLSK-Net improves as the receptive field of the large-kernel convolution group increases.

\begin{table}[h]
\renewcommand\arraystretch{1.2}
\vspace{-1\baselineskip}
\caption{The impact of different configuration selection mechanisms on the downscaling performance of MoCoLSK-Net. S: MoCoLSK-SS, C: MoCoLSK-CS. For example, (C, S, C, C) in second column means that second stage of MoCoLSK-Net uses MoCoLSK and other stages use MoCoLSK-CS.}\label{tab:7-selection}
\vspace{2pt}
\centering
\setlength{\tabcolsep}{3.5pt}

\begin{tabular}{c|c|cc}
Fusion Modules     & S/C Sequences & RMSE$\downarrow$            & CC$\uparrow$              \\ \midrule[1pt]
MoCoLSK-SS only    & (S, S, S, S)         & 0.7133          & 0.9613          \\
MoCoLSK-CS only & (C, C, C, C)         & 0.7486          & 0.9599          \\ \hline
\multirow{6}{*}{\begin{tabular}[c]{@{}c@{}}Interleaved \\ MoCoLSK-SS \& \\ MoCoLSK-CS\end{tabular}} & (S, C, S, C) & 0.7044 & 0.9627 \\
           & (C, S, C, S)         & 0.7040          & 0.9627          \\
           & (C, C, S, S)         & 0.7181          & 0.9609          \\
           & (S, S, C, C)         & 0.7077          & 0.9621          \\
           & (C, S, C, C)         & \textbf{0.7000} & \textbf{0.9633} \\
           & (S, C, S, S)         & 0.7205          & 0.9606         
\end{tabular}
\end{table}

\subsection{Spatial Selection or Channel Selection?}

LSKNet \cite{ICCV2023LSK} introduces both the large spatially selective kernel module (LSK-SS) and the large channel-selective kernel module (LSK-CS). Based on this, we develop MoCoLSK-CS module, integrating modality-conditioned weights generated from MCWG pathway with features obtained from LSK-CS after global average pooling, using element-wise addition.

The default configuration of MoCoLSK-Net comprises four stages, each containing a MoCoLSK module (referred to as MoCoLSK-SS). To investigate which selection mechanism is more effective, we configure different selection mechanisms for the four stages: MoCoLSK-SS for spatial selection or MoCoLSK-CS for channel selection. Table \ref{tab:7-selection} presents all selection mechanism configurations and their corresponding experimental results, leading to the following conclusions:
reconstruction performance of MoCoLSK-Net configured exclusively with MoCoLSK-SS is superior to that of the network configured solely with MoCoLSK-CS modules. This indicates that spatial selection is significantly more critical than channel selection for LST downscaling tasks. 
Moreover, interleaving MoCoLSK-SS and MoCoLSK-CS modules within MoCoLSK-Net achieves relatively better reconstruction performance compared to configurations using only MoCoLSK-SS or MoCoLSK-CS. For instance, configurations such as (S, C, S, C), (C, S, C, S), (S, S, C, C), and especially (C, S, C, C), support this observation.

\begin{figure}[t]
    \centering
    \includegraphics[width=.8\columnwidth]{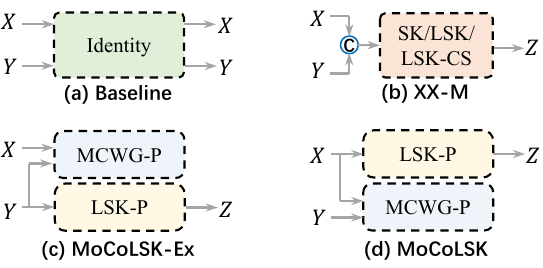}
    \caption{
    Thumbnails of different multimodal selection mechanisms (up/down-projection layers are ignored). $\bm{X}$: HR LST, $\bm{Y}$: HR guidance, P: pathway, M: multimodal.  (a) Baseline, meaning the output is the same as the input. (b) XX-M represents three modules: SK-M, LSK-M, and LSK-CS-M, which concatenate $\bm{X}$ and $\bm{Y}$ before feeding them into the original SK \cite{CVPR2019SKNet}, LSK \cite{ICCV2023LSK}, and LSK-CS \cite{ICCV2023LSK} modules. (c) MoCoLSK-Ex (pathway exchange) means the guidance feature $\bm{Y}$ enters the LSK pathway, and the output of modality fusion is denoted as $\bm{Z}= \bm{X} \otimes \bm{S}$. (d) Our MoCoLSK module.
    }
    \label{fig:multimodal-sm}
    % \vspace{-1\baselineskip}
\end{figure}

\subsection{Comparison of Different Multimodal Selective Mechanisms.}

To further explore the effectiveness of our MoCoLSK module, we compare it with several other multimodal variants. Fig. \ref{fig:multimodal-sm} presents the schematic diagrams and configurations of all the variants. The conclusions from Table \ref{tab:8-multimodal} are as follows:

1) The SK-M and LSK-CS-M modules demonstrate poor reconstruction performance, reaffirming that a multimodal fusion strategy relying solely on channel selection mechanisms may not be suitable for LST downscaling tasks.

2) The LSK-M module exhibits excellent reconstruction performance, nearly matching that of our MoCoLSK module. This highlights the importance of the spatial selection mechanism, especially for the challenges posed by LST downscaling.

3) The MoCoLSK-Ex module performs worse than our MoCoLSK module and even falls below the baseline. This indicates that feeding LST features, rather than guidance features, into the LSK pathway is critical for  reconstruction.

\begin{table}[t]
\renewcommand\arraystretch{1.2}
\vspace{-1\baselineskip}
\caption{Comparative study of different multimodal selective mechanisms.}\label{tab:8-multimodal}
\vspace{2pt}
\centering
\setlength{\tabcolsep}{6pt}

\begin{tabular}{c|c|cc}
No. & Fusion Modules & RMSE$\downarrow$            & CC$\uparrow$              \\ \midrule[1pt]
1   & Baseline       & 0.7405          & 0.9605          \\
2   & SK-M           & 1.0700          & 0.9181          \\
3   & LSK-M          & 0.7193          & 0.9612          \\
4   & LSK-CS-M       & 0.7314          & 0.9610          \\
5   & MoCoLSK-Ex     & 0.7461          & 0.9587          \\
6   & MoCoLSK        & \textbf{0.7133} & \textbf{0.9613}    
\end{tabular}
\vspace{-1\baselineskip}
\end{table}

\begin{figure}[htbp]
    \centering
    \includegraphics[width=.9\columnwidth]{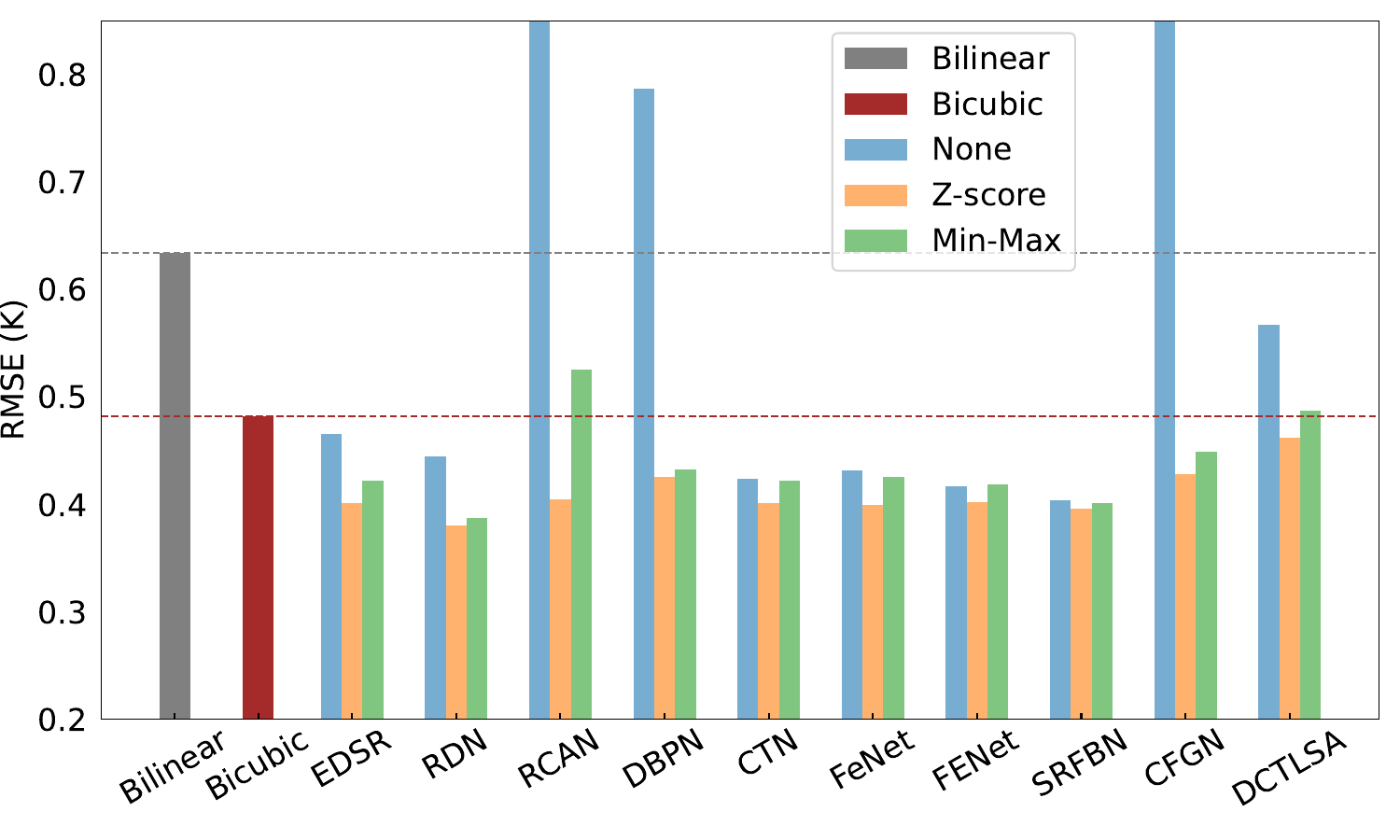}
    \caption{RMSE comparison between existing SOTA SISR methods ($\times$2) under different normalization strategies. The experiments follow the default configuration.}
    \label{fig:norm-sisr}
    \vspace{-1\baselineskip}
\end{figure}

\begin{figure}[htbp]
    \centering
    \includegraphics[width=.9\columnwidth]{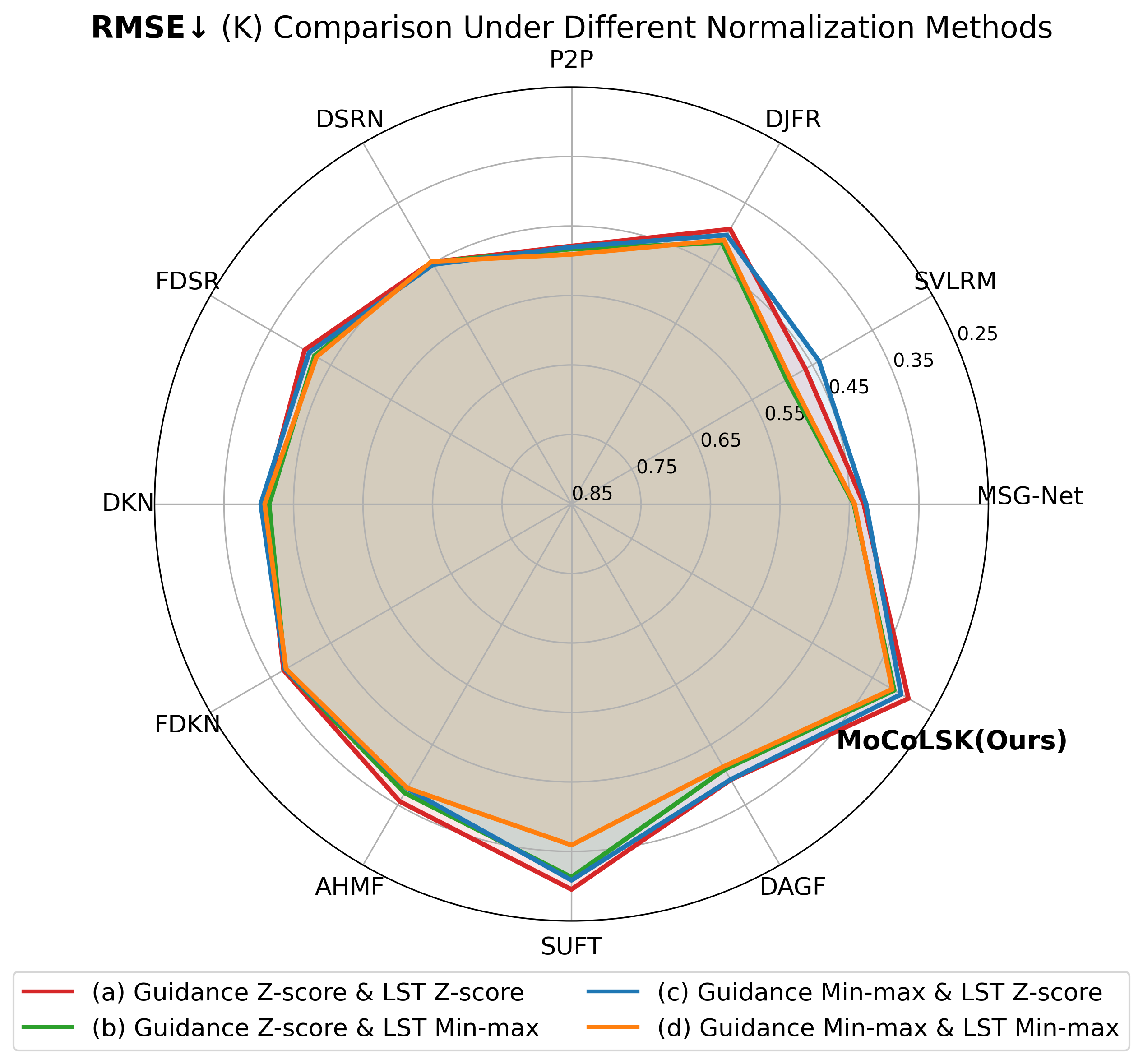}
    \caption{RMSE comparison between our MoCoLSK-Net and existing SOTA GDSR methods ($\times$2) under different normalization strategies. The experiments follow the default configuration.}
    % \caption{Test Taylor diagram.}
    \label{fig:norm-gdsr-radar}
    \vspace{-1\baselineskip}
\end{figure}

\begin{table*}[htbp]
  \renewcommand\arraystretch{1.2}
  \footnotesize
  \centering
  \vspace{-1\baselineskip}
  \caption{Comparison of Various Loss Functions on MoCoLSK Reconstruction Performance.}\label{tab:loss}
  \vspace{-2pt}
  \setlength{\tabcolsep}{2pt}
  
  \begin{tabular}{l|ccccc|ccccc|ccccc}
  \multirow{2}{*}{Loss} & \multicolumn{5}{c|}{$\times 2$} & \multicolumn{5}{c|}{$\times 4$} & \multicolumn{5}{c}{$\times 8$} \\
  & RMSE$\downarrow$ & MAE$\downarrow$ & BIAS & CC$\uparrow$ & RSD$\downarrow$ 
  & RMSE$\downarrow$ & MAE$\downarrow$ & BIAS & CC$\uparrow$ & RSD$\downarrow$ 
  & RMSE$\downarrow$ & MAE$\downarrow$ & BIAS & CC$\uparrow$ & RSD$\downarrow$  \\
  \Xhline{1pt}
    L1 & 0.2902 & 0.1951 & 0.0009 & 0.9937 & 0.0062 & 0.5590 & 0.3883 & 0.0020 & 0.9771 & 0.0218 & 0.8031 & 0.5642 & 0.0027 & 0.9514 & 0.0456 \\
    SSIM & 0.2932 & 0.2011 & 0.0112 & 0.9938 & 0.0052 & 0.5668 & 0.4015 & 0.0096 & 0.9770 & 0.0184 & 0.8388 & 0.6052 & 0.0138 & 0.9479 & 0.0422 \\
    MS-SSIM & 0.2935 & 0.2009 & 0.0000 & 0.9937 & 0.0059 & 0.5745 & 0.4058 & 0.0003 & 0.9763 & 0.0215 & 0.8229 & 0.5908 & 0.0075 & 0.9504 & 0.0429 \\
    0.3 SSIM + 0.7 L1 & 0.2906 & 0.1956 & 0.0011 & 0.9937 & 0.0063 & 0.5606 & 0.3899 & 0.0017 & 0.9771 & 0.0219 & 0.8030 & 0.5647 & 0.0024 & 0.9518 & 0.0451 \\
    0.5 SSIM + 0.5 L1 & 0.2898 & 0.1950 & 0.0010 & 0.9938 & 0.0062 & 0.5642 & 0.3925 & 0.0020 & 0.9769 & 0.0221 & 0.7982 & 0.5615 & 0.0034 & 0.9521 & 0.0446 \\
    0.7 SSIM + 0.3 L1 & 0.2886 & 0.1943 & 0.0009 & 0.9938 & 0.0061 & 0.5587 & 0.3889 & 0.0020 & 0.9771 & 0.0217 & 0.8028 & 0.5658 & 0.5658 & 0.9514 & 0.0456 \\
    0.84 SSIM + 0.16 L1 & 0.2890 & 0.1948 & 0.0009 & 0.9938 & 0.0061 & 0.5679 & 0.3956 & 0.0019 & 0.9765 & 0.0225 & 0.8033 & 0.5673 & 0.0031 & 0.9514 & 0.0453 \\
    0.84 MS-SSIM + 0.16 L1 & 0.2890 & 0.1945 & 0.0011 & 0.9938 & 0.0061 & 0.5614 & 0.3907 & 0.0017 & 0.9770 & 0.0221 & 0.8104 & 0.5700 & 0.0026 & 0.9506 & 0.0465
  \end{tabular}
  \vspace{-1\baselineskip}
\end{table*}

\subsection{Which Normalization Performs Best?}\label{subsec:norm}
% The guided LST downscaling task comprises two parts of data: the LST data to be super-resolved and the guidance data. 
% We raise two crucial questions: 
% \emph{1) Is it necessary to normalize both modalities of data? }
% \emph{2) Which normalization method should be adopted for both types of data to maximize the model's performance?}

% We investigated three normalization techniques on a large number of SOTA deep learning-based downscaling methods, including single image downscaling and guided image downscaling methods: no normalization (denoted as None), Z-score, and Min-Max.
We delve into the impact of three normalization strategies on the performance of a downscaling method: no normalization (denoted as None), Z-score, and Min-max.
The definitions of Z-score and Min-max are as follows:
\begin{equation}
\text{Z-score}(X)=\frac{X-\bar{\mu}}{\bar{\sigma}} ,
\end{equation}
\begin{equation}
% \text{Min-max}(X)=\frac{X-X_{\min}}{X_{\max}-X_{\min}},
\text{Min-max}(X)=\frac{X-\bar{X}_{min} }{\bar{X}_{max} -\bar{X}_{min} },
\end{equation}
where $X$ is LST data, $\bar{\mu}$, $\bar{\sigma}$, $\bar{X}_{min}$ and $\bar{X}_{max}$ are the mean, standard deviation, minimum value, and maximum value of all LST data in GrokLST dataset, respectively.

Fig. \ref{fig:norm-sisr} compares the RMSE of ten single-image super-resolution methods applied to LST data using different normalization strategies. It is evident that the None strategy performs significantly worse than the Z-score and Min-max strategies, with Z-score achieving the best results. This strongly highlights the necessity of normalization.
Fig. \ref{fig:norm-gdsr-radar} presents the RMSE results of twelve guided downscaling methods applied to LST data and guidance data using different normalization strategies. It can be observed that the Z-score strategy is particularly suitable for LST data. For example, models employing strategies (a) and (c) exhibit significantly better reconstruction performance compared to those using strategies (b) and (d). 
Furthermore, models utilizing strategies (a) and (c) achieve commendable reconstruction performance, with strategy (a) yielding the most superior results. This indicates that the Z-score strategy is effective not only for LST data but also for guidance data.
In conclusion, we believe that normalization strategies are essential for downscaling tasks. Whether for single-image downscaling or guided downscaling methods, the Z-score strategy is worth considering.

\subsection{Which Loss Works Best?}

To explore the impact of different loss functions on MoCoLSK reconstruction performance, we conduct a detailed comparison using several mainstream loss functions, including L1 loss, structural similarity index measure (SSIM) loss, multi scale structural similarity index measure (MS-SSIM) loss, and their combinations. Key insights from Table \ref{tab:loss} are as follows:

1) Using SSIM or MS-SSIM alone as the loss function to supervise MoCoLSK-Net learning yielded noticeably worse performance than using L1 loss alone, which was observed across all three reconstruction scales.

2) When combining SSIM and L1 loss, MoCoLSK-Net achieved the best reconstruction metrics across all three scales. Furthermore, increasing the weight of the SSIM loss slightly improved reconstruction results, though the improvements were minimal.

3) When combining MS-SSIM and L1 loss, we observed a slight improvement in the $\times$2 reconstruction task compared to L1 loss alone. However, at larger scales (i.e., $\times$4, $\times$8), we saw the opposite trend, with performance even worse than when using MS-SSIM alone.

Overall, we recommend using a combination of SSIM loss with a higher weight and L1 loss with a lower weight for LST downscaling research.

% !TEX root = ../main.tex

\section{Conclusion} \label{sec:conclusion}

To promote the thriving development of LST downscaling, we contribute a comprehensive open-source ecosystem, GrokLST project, which includes GrokLST dataset, a high-resolution benchmark dataset specifically designed for LST downscaling, and a toolkit featuring over 40 advanced downscaling methods along with various downscaling metrics. Additionally, we propose a novel and effective modality-conditioned multimodal fusion network, MoCoLSK-Net, to address guided LST downscaling challenges. Through extensive quantitative and qualitative comparisons on GrokLST dataset with four machine learning methods, nineteen single-image downscaling methods, and thirteen guided image downscaling methods, MoCoLSK-Net demonstrates superior reconstruction performance, achieving the most accurate LST predictions.

% use section* for acknowledgment
\section*{Acknowledgment}

The authors would like to thank the International Research Center of Big Data for Sustainable Development Goals (CBAS) for kindly providing the SDGSAT-1 data. We acknowledge the Tianjin Key Laboratory of Visual Computing and Intelligent Perception (VCIP) for their essential resources. Computation is partially supported by the Supercomputing Center of Nankai University (NKSC).

\bibliographystyle{IEEEtran}
\bibliography{./reference.bib}

% !TEX root = ../main.tex

% Qun Dai
\begin{IEEEbiography}[{\includegraphics[width=1in,height=1.25in,clip,keepaspectratio]{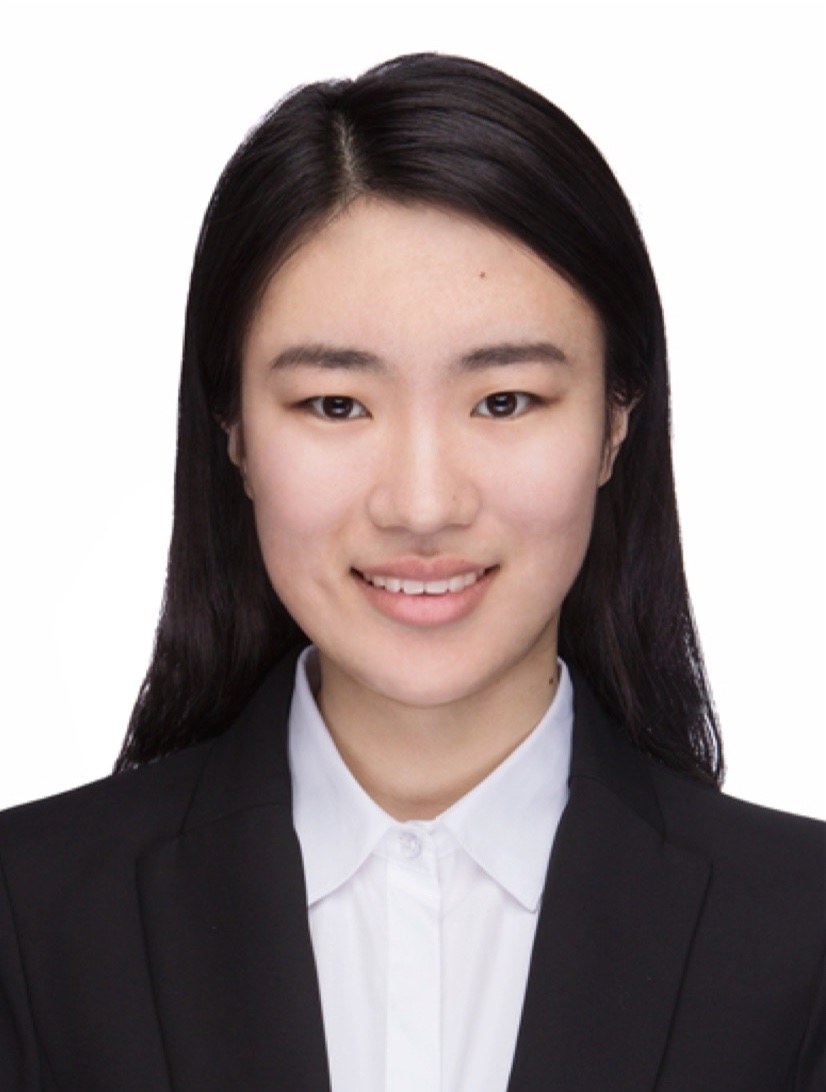}}]{Qun Dai} received her M.S. degree in Avionics Engineering in 2017. Subsequently, she worked as an Avionics Engineer at China Eastern Airlines Technology Co., Ltd., where she gained extensive experience in the field of aviation electronics and systems integration. 
Currently, she is a Ph.D. candidate at Nanjing University of Science and Technology, where she focuses on advancing the state-of-the-art in image restoration techniques for enhanced object detection performance. 
\end{IEEEbiography}

% Chunyang Yuan
\begin{IEEEbiography}[{\includegraphics[width=1in,height=1.25in,clip,keepaspectratio]{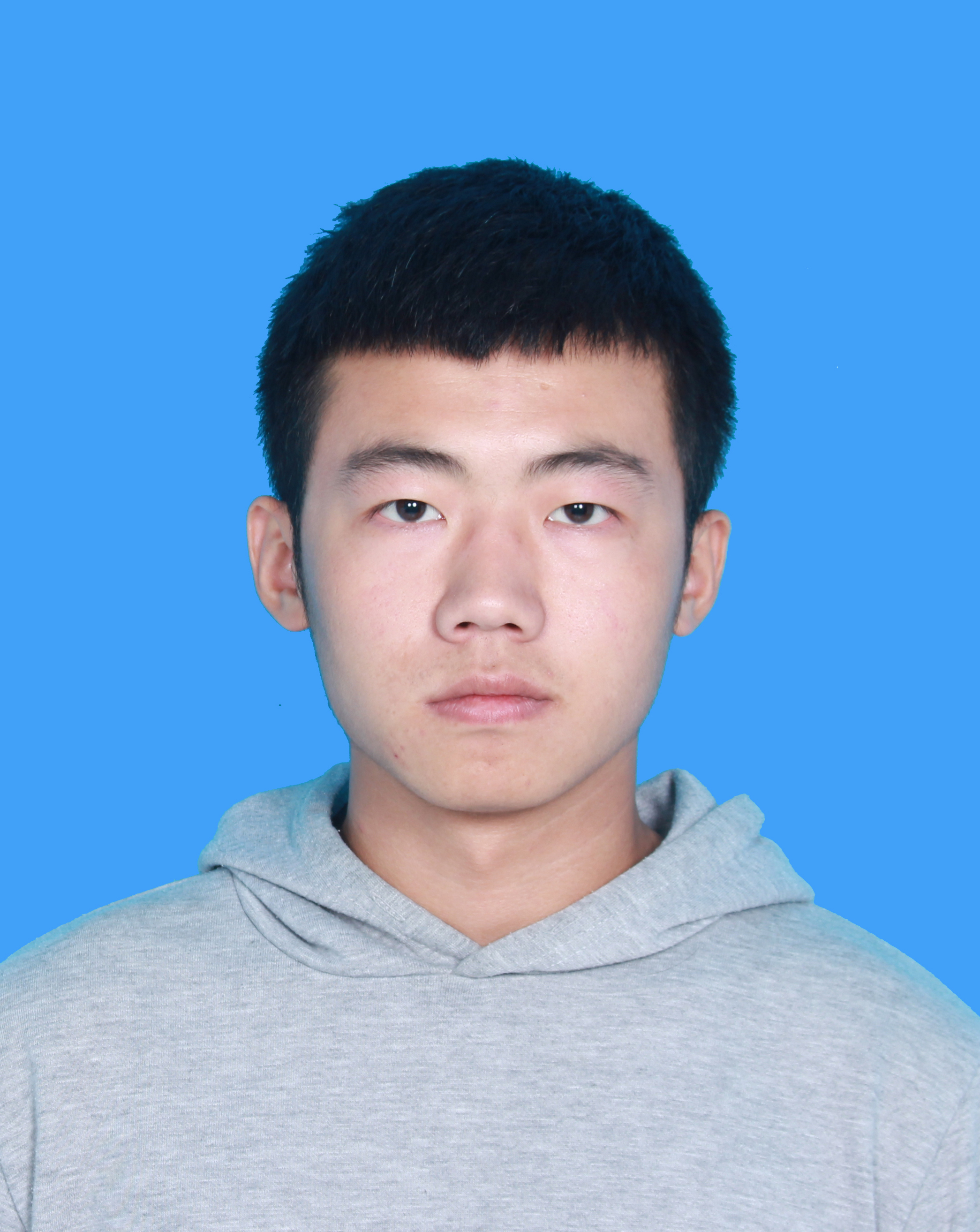}}]{Chunyang Yuan} received the B.S degree in energy and power engineering from Henan Polytechnic University, Henan, China, in 2021. He is currently pursuing a master's degree in Computer Science and Technology at Nanjing University of Posts and Telecommunications, Nanjing, China. His research interests include machine learning, SAR image processing, and computer vision.
\end{IEEEbiography}

% Yimian Dai
\begin{IEEEbiography}[{\includegraphics[width=1in,height=1.25in,clip,keepaspectratio]{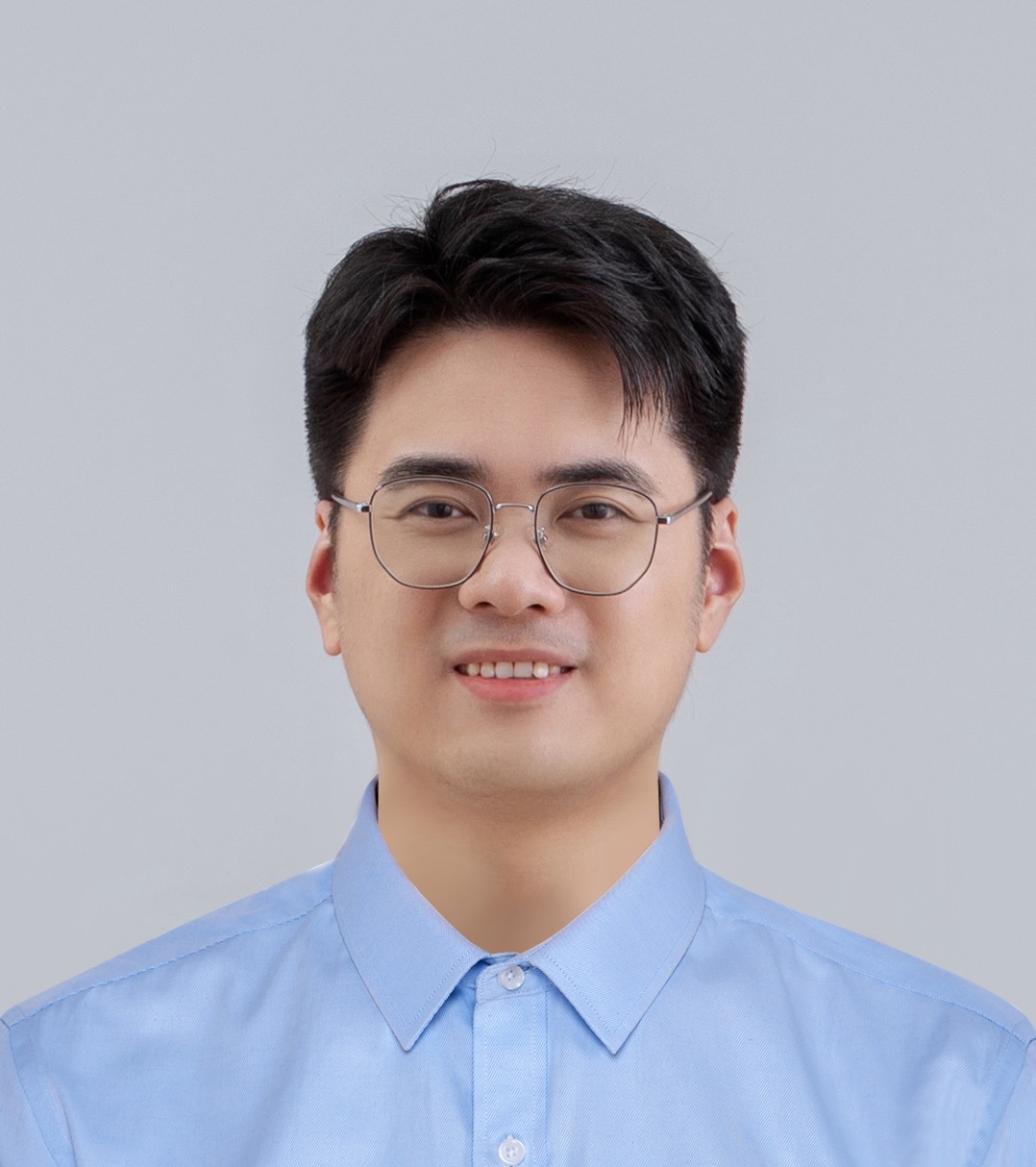}}]{Yimian Dai} 
(Member, IEEE) received the B.E. degree in information engineering and the Ph.D. degree in signal and information processing from Nanjing University of Aeronautics and Astronautics, Nanjing, China, in 2013 and 2020, respectively.
From 2021 to 2024, he was a Postdoctoral Researcher with the School of Computer Science and Engineering, Nanjing University of Science and Technology, Nanjing, China. 
He is currently an Associate Professor with the College of Computer Science, Nankai University, Tianjin, China.
His research interests include computer vision, deep learning, and their applications in remote sensing.
For more information, please visit the link (\href{https://yimian.grokcv.ai/}{https://yimian.grokcv.ai/}).

\end{IEEEbiography}

% Yuxuan Li
\begin{IEEEbiography}[{\includegraphics[width=1.1in,height=1.35in,clip,keepaspectratio]{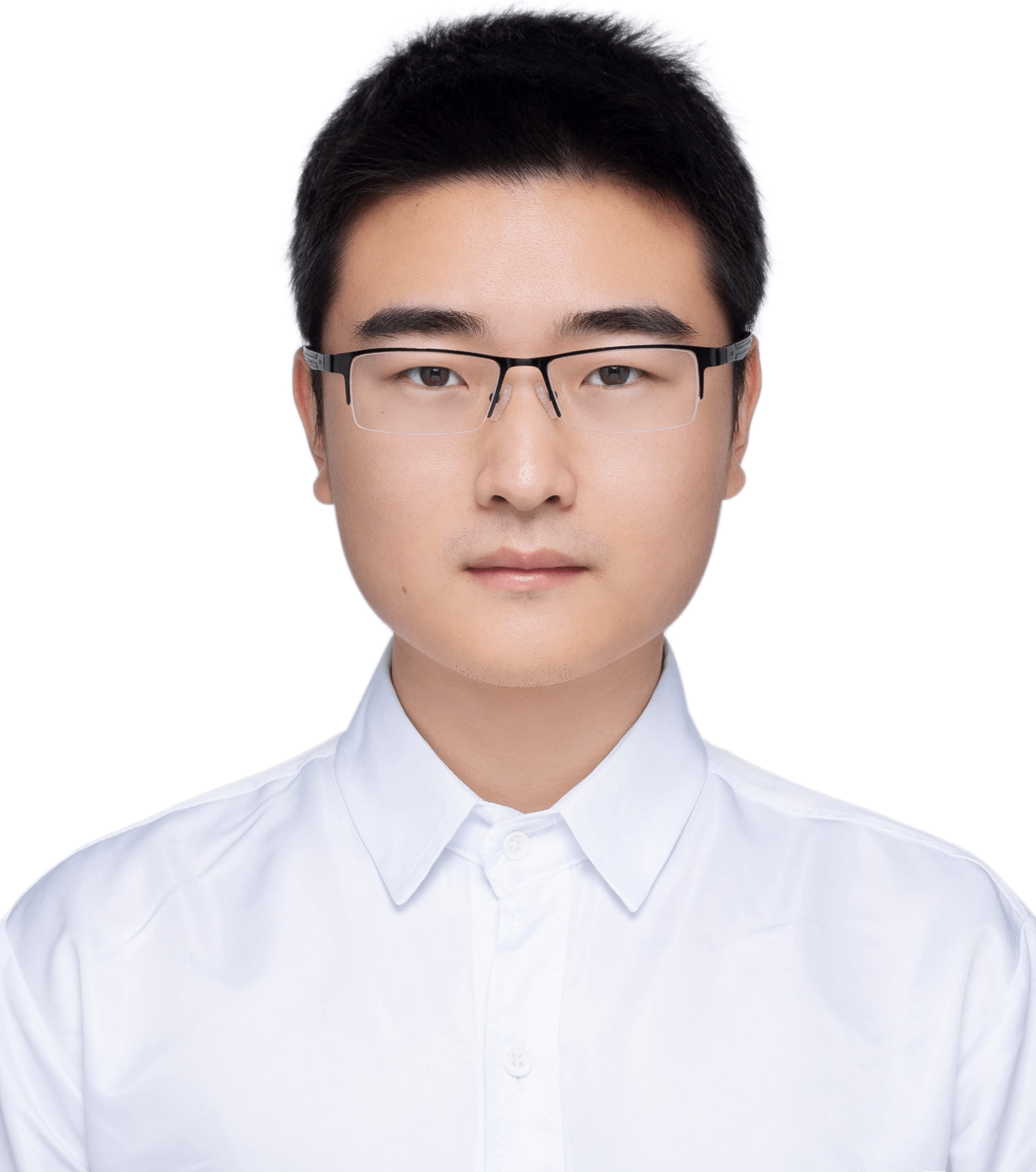}}]{Yuxuan Li} is currently a Ph.D. student at the Department of Computer Science, Nankai University, China.
  He graduated from University College London (UCL) with a first-class degree in Computer Science. He was the champion of the Second Jittor Artificial Intelligence Challenge in 2022, was awarded 2nd place in Facebook Hack-a-Project in 2019 and was awarded 2nd place in the Greater Bay Area International Algorithm Competition in 2022. His research interests include neural architecture design, and remote sensing object detection. 
\end{IEEEbiography}

% Xiang Li
\begin{IEEEbiography}[{\includegraphics[width=1.1in,height=1.35in,clip,keepaspectratio]{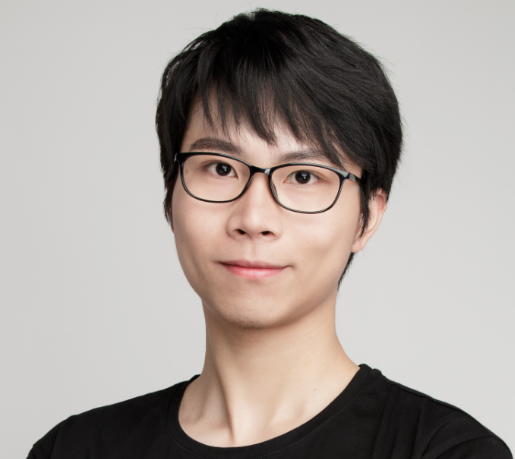}}]{Xiang Li} is an Associate Professor in College of Computer Science, Nankai University. He received the PhD degree from the Department of Computer Science and Technology, Nanjing University of Science and Technology (NJUST) in 2020. There, he started the postdoctoral career in NJUST as a candidate for the 2020 Postdoctoral Innovative Talent Program. In 2016, he spent 8 months as a research intern in Microsoft Research Asia, supervised by Prof. Tao Qin and Prof. Tie-Yan Liu. He was a visiting scholar at Momenta, mainly focusing on monocular perception algorithm. His recent works are mainly on: neural architecture design, CNN/Transformer, object detection/recognition, unsupervised learning, and knowledge distillation. He has published 20+ papers in top journals and conferences such as TPAMI, CVPR, NeurIPS, etc.
\end{IEEEbiography}

% 周霞
% \begin{IEEEbiography}[{\includegraphics[width=1in,height=1.25in,clip,keepaspectratio]{./figs/biography/Xia_Zhou}}]{Xia Zhou} received the M.S. and Ph.D. degrees from the South China Botanical Garden, Chinese Academy of Sciences and University of Chinese Academy of Sciences, China, in 2002 and 2009, respectively.
% She is currently a director of the Guangzhou Institute of Geography, Guangdong Academy of Sciences, Guangzhou, China. Her main research interests include remote sensing and applications on ecological environment, and biodiversity. She has published more than 30 articles in prestigious journals. Her work has been funded by multiple grants from the National Natural Science Foundation of China, the Guangdong Natural Science Foundation, and the Guangdong Science and Technology Plan Project.
% \end{IEEEbiography}

% Kang Ni
\begin{IEEEbiography}[{\includegraphics[width=1.1in,height=1.35in,clip,keepaspectratio]{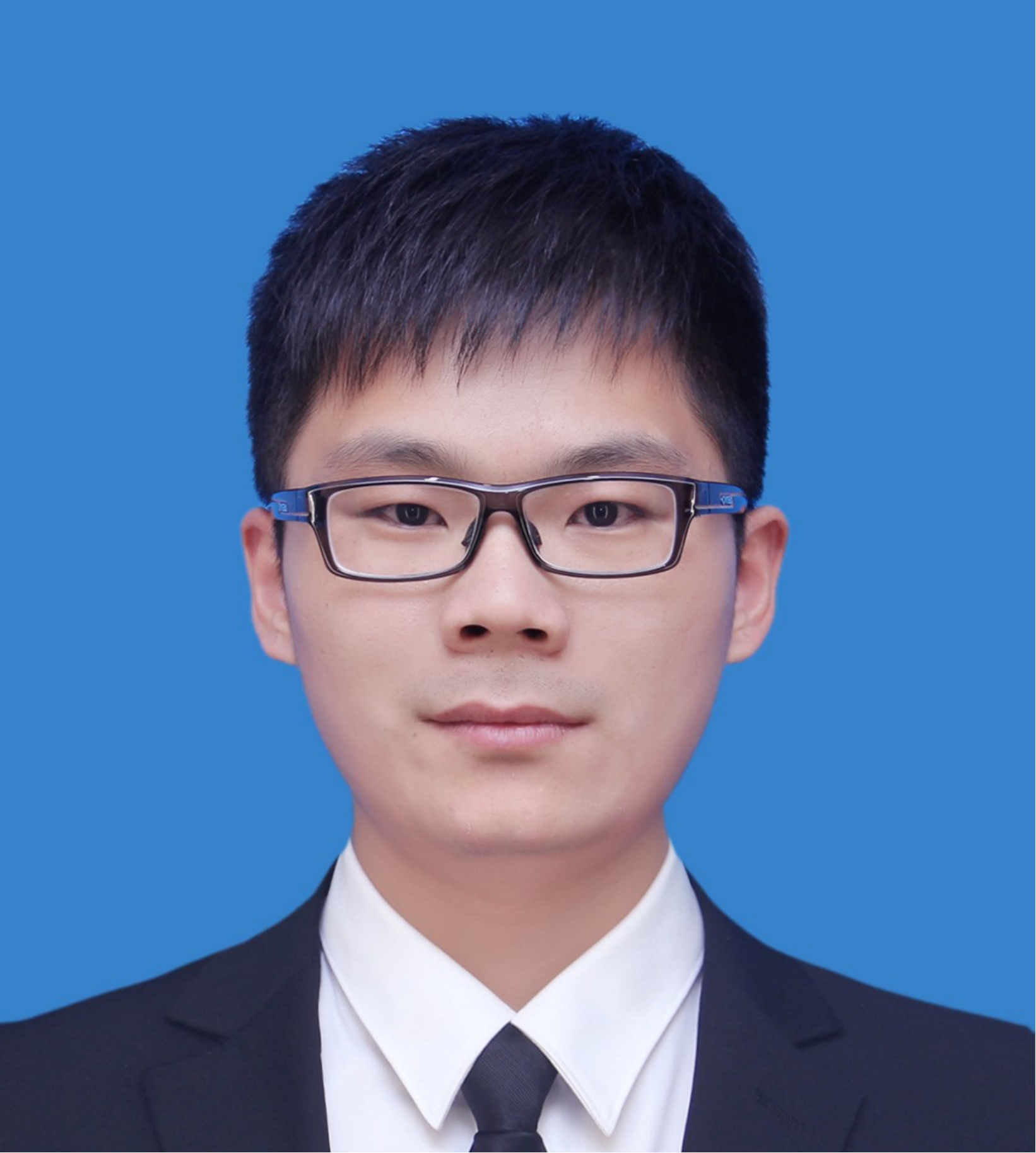}}]{Kang Ni (Member, IEEE) } received the M.S. degrees from Changchun University of Technology, Jilin, China, in 2016, and the Ph.D. degree from Nanjing University of Aeronautics and Astronautics, Jiangsu, China, in 2020.
He is an Associate Professor with the School of Computer Science, Nanjing University of Posts and Telecommunications, Nanjing, and also a member with the Jiangsu Key Laboratory of Big Data Security and Intelligent Processing, Nanjing. His research interests include machine learning, SAR image processing, and computer vision.
\end{IEEEbiography}

% 许剑辉
\begin{IEEEbiography}[{\includegraphics[width=1in,height=1.25in,clip,keepaspectratio]{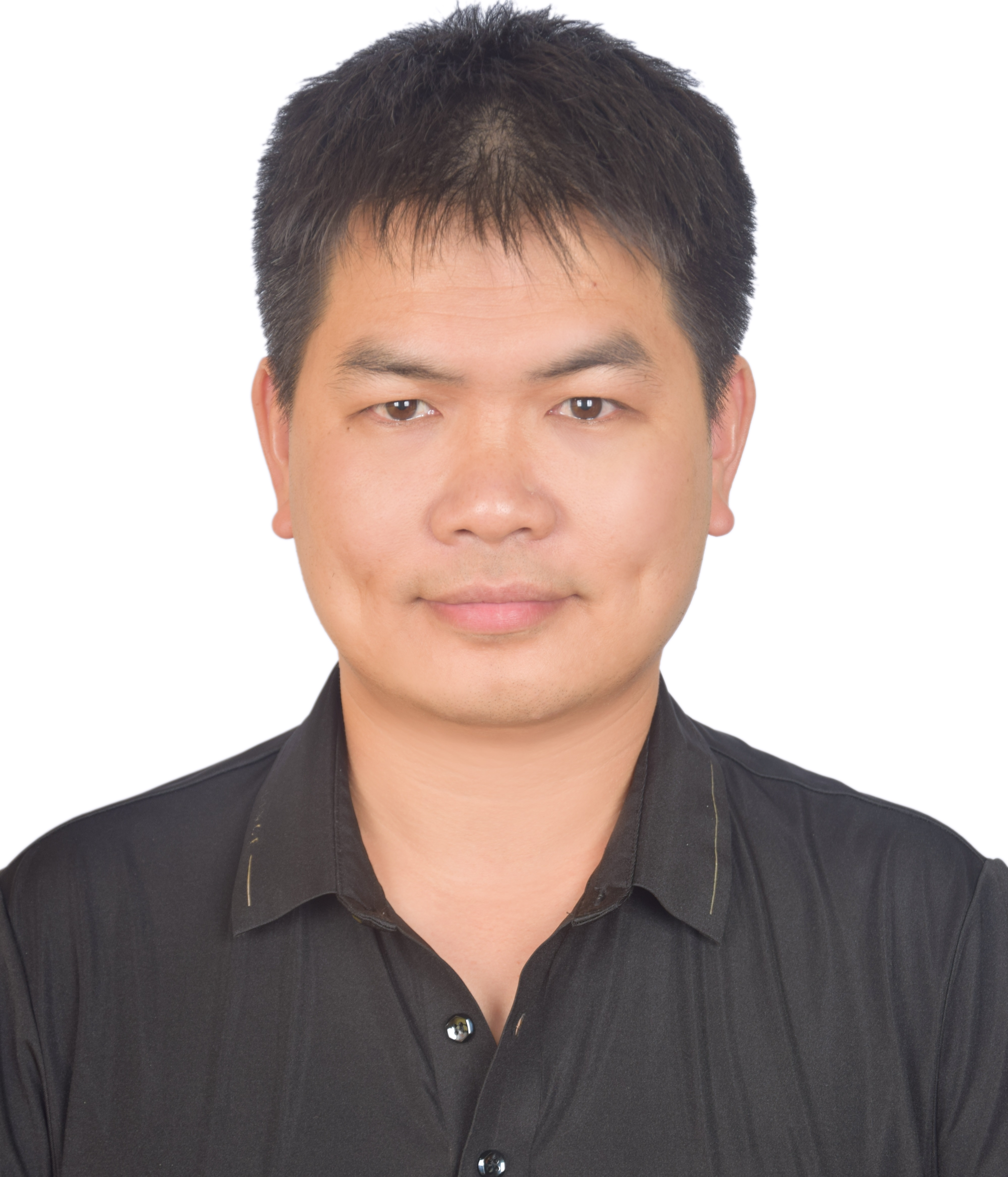}}]{Jianhui Xu} received the Ph.D. degree in multi-source remote sensing data assimilation from Wuhan University, Wuhan, China, in 2015. He is currently an Associate Professor with the Guangzhou Institute of Geography, Guangdong Academy of Sciences. He has published numerous SCI papers in prestigious journals, such as The Science of The Total Environment, Journal of Geophysical Research Atmospheres and Building and Environment, in the area of data fusion and urban remote sensing. His research interests include data fusion and assimilation, land surface temperature and urban remote sensing.
\end{IEEEbiography}

% Xiangbo Shu
\begin{IEEEbiography}[{
  \includegraphics[width=1.45in,height=1.3in,clip,keepaspectratio]{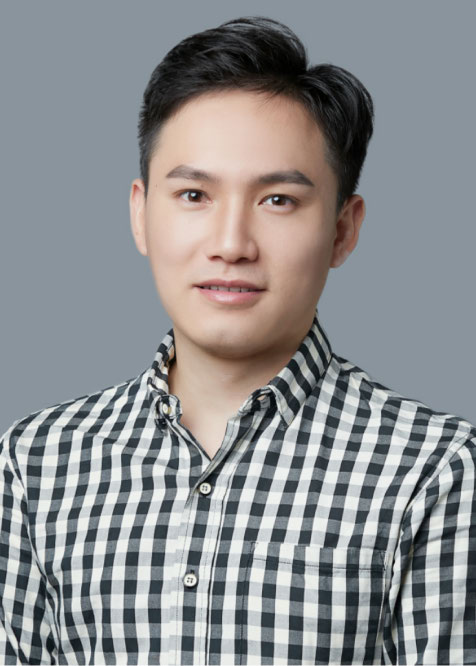}}]{Xiangbo Shu} (Senior Member, IEEE) received the Ph.D. degree from Nanjing University of Science and Technology, Nanjing, China, in 2016.

From 2014 to 2015, he worked as a Visiting Scholar at the National University of Singapore, Singapore. He is currently a Professor with the School of Computer Science and Engineering, Nanjing University of Science and Technology. His current research interests include computer vision, and multimedia. He has authored over 100 journals and conference papers in these areas.
% , including IEEE TRANSACTIONS ON PATTERN ANALYSIS AND MACHINE INTELLIGENCE (TPAMI), TRANSACTIONS ON NEURAL NETWORKS AND LEARNING SYSTEMS (TNNLS), IEEE TRANSACTIONS ON IMAGE PROCESSING (TIP), IEEE Conference on Computer Vision and Pattern Recognition (CVPR), IEEE International Conference on Computer Vision (ICCV), European Conference on Computer Vision (ECCV), and ACM International Conference on Multimedia (ACM MM).

Dr. Shu is a member of ACM and a Senior Member of CCF. He has received the Best Student Paper Award in International Conference on MultiMedia Modeling (MMM) 2016 and the Best Paper Runner-Up in ACM MM 2015. He has served on an editorial boards for IEEE TNNLS, IEEE TRANSACTIONS ON CIRCUITS AND SYSTEMS FOR VIDEO (TCSVT), and Information Sciences.
\end{IEEEbiography}

% Jian Yang
\begin{IEEEbiography}[{
  \includegraphics[width=1.45in,height=1.3in,clip,keepaspectratio]{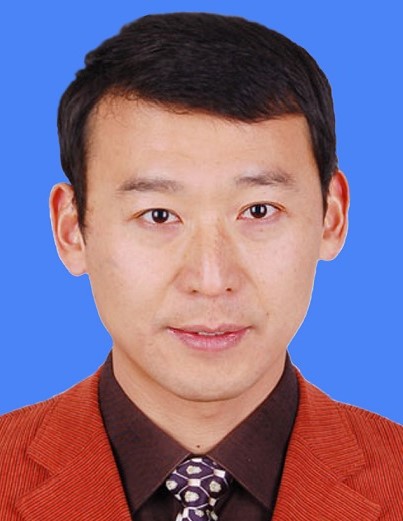}}]{Jian Yang} received the PhD degree from Nanjing University of Science and Technology (NJUST) in 2002, majoring in pattern recognition and intelligence systems. From 2003 to 2007, he was a Postdoctoral Fellow at the University of Zaragoza, Hong Kong Polytechnic University and New Jersey Institute of Technology, respectively. From 2007 to present, he is a professor in the School of Computer Science and Technology of NJUST. His papers have been cited over 50000 times in the Scholar Google. His research interests include pattern recognition and computer vision. Currently, he is/was an associate editor of Pattern Recognition, Pattern Recognition Letters, IEEE Trans. Neural Networks and Learning Systems, and Neurocomputing. He is a Fellow of IAPR. 
\end{IEEEbiography} % 作者介绍

\end{document}